\documentclass[11pt]{article}

\usepackage[preprint]{acl}

\usepackage{times}
\usepackage{latexsym}

\usepackage[T1]{fontenc}

\usepackage[utf8]{inputenc}

\usepackage{microtype}

\usepackage{inconsolata}

\usepackage{graphicx}
\usepackage{subcaption}
\usepackage{amsmath}
\usepackage{amssymb}
\usepackage{booktabs}
\usepackage{multirow}
\usepackage{xcolor}
\usepackage{enumitem}
\usepackage{algorithm}
\usepackage[noend]{algpseudocode}
\usepackage{mdframed}
\mdfdefinestyle{grayquote}{
  backgroundcolor=gray!12,
  linewidth=0pt,
  leftmargin=1.5em,
  rightmargin=1.5em,
  innerleftmargin=0.6em,
  innerrightmargin=0.6em,
  innertopmargin=0.35em,
  innerbottommargin=0.35em,
  skipabove=0.4em,
  skipbelow=0.4em,
}
\newenvironment{grayquote}{\begin{mdframed}[style=grayquote]\small}{\end{mdframed}}
\title{The Illusion of Visual Tool-Use: A Causal Audit of Thinking with Images}

\author{
 \textbf{Zhiheng Wang\textsuperscript{1,2}},
 \textbf{Bo Peng\textsuperscript{1,2,3}},
 \textbf{Lai Wei\textsuperscript{1}},
 \textbf{Chaochao Lu\textsuperscript{2}\thanks{Corresponding author.}}
\\
 \textsuperscript{1}Shanghai Jiao Tong University\\
 \textsuperscript{2}Shanghai Artificial Intelligence Laboratory\\
 \textsuperscript{3}Shanghai Innovation Institute
\\
{ 
    \texttt{\{wangzhiheng,pengbo,luchaochao\}@pjlab.org.cn}
 }
}

\begin{document}
\maketitle
\begin{abstract}

The ``thinking-with-images'' paradigm equips multimodal LLMs with active visual operations such as crop-and-zoom. However, models using these operations often achieve only marginal or negative gains over direct inference at substantially higher token cost. They may also repeatedly crop irrelevant regions and fail on questions that direct inference answers correctly. We ask \textbf{whether the returned visual evidence causally affects the answer}. To answer this question, we formulate visual tool-use as a causal graph that separates \emph{observation-mediated paths} from \emph{action-induced shortcuts}. We then audit it through interventions at the three levels: policy (comparing tool-use with direct inference), trajectory (corrupting all observations during rollout), and step (counterfactually replacing one individual observation under a fixed prefix). Our step-level estimand, \emph{Visual Evidence Gain}, isolates the contribution of each returned observation. Across six representative models and five fine-grained perception benchmarks, we uncover \emph{policy miscalibration} with two failure modes. In \emph{Calling Without Looking}, returned observations have no causal effect on the answer. In \emph{Looking Without Planning}, observations are informative but the call schedule is incoherent. A trajectory-level diagnostic decomposes the policy-level accuracy gain into per-group contributions and shows that the gain is concentrated in a \emph{Calibrated} minority. We term this discrepancy the \textbf{\emph{illusion of visual tool-use}}: despite aggregate accuracy gains, visual tool-use is not causally effective across a broad range of rollouts. The code is available at \url{https://github.com/OpenCausaLab/CauAudit}.
\end{abstract}

\section{Introduction}

The recent evolution of Multimodal Large Language Models (MLLMs) \citep{openai2025o3o4mini,bai2025qwen3vl} has been increasingly driven by the ``thinking-with-images'' paradigm~\citep{openai2025thinkingwithimages,zhang2025causight,zheng2025deepeyes,wang2025pixel,lai2025mini-o3}, which interleaves reasoning with visual operations such as crop-and-zoom. This enables the model to gather fine-grained evidence beyond a single holistic view.

\begin{figure}[!t]
  \centering
  \begin{subfigure}[b]{\columnwidth}
      \centering
      \includegraphics[width=\columnwidth]{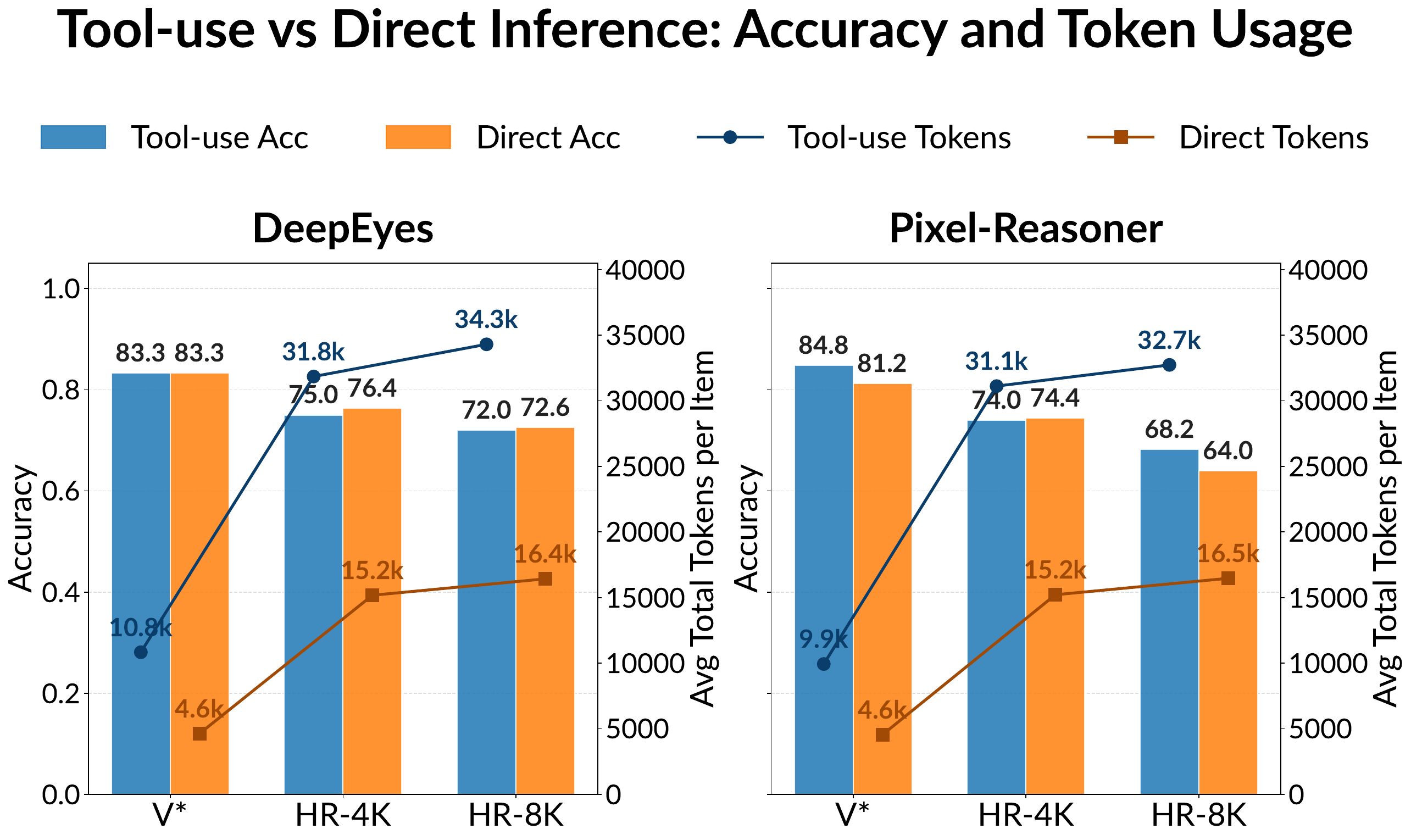}
      \caption{The Marginal Improvement Paradox}
      \label{fig:teaser_a}
  \end{subfigure}
  \\[0.6em]
  \begin{subfigure}[b]{\columnwidth}
      \centering
      \includegraphics[width=\columnwidth]{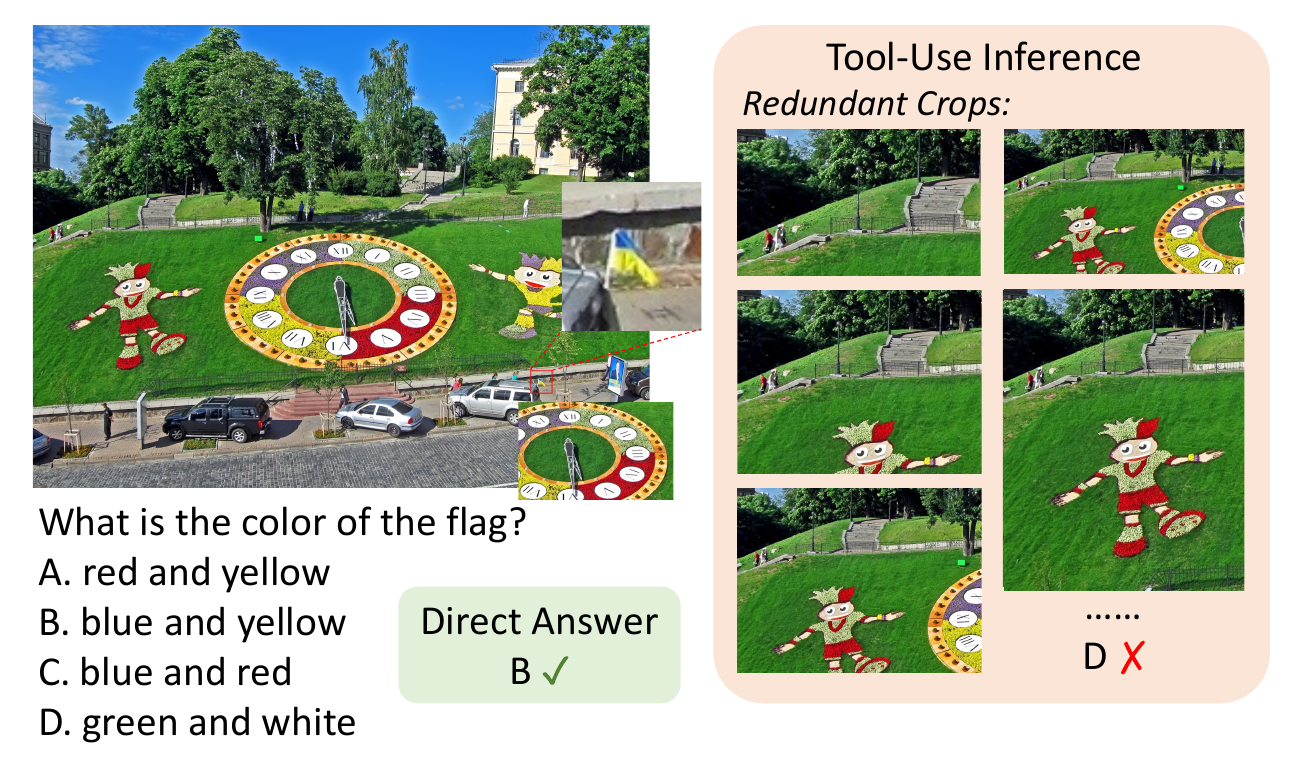}
      \caption{The Zoom-Loop Trap}
      \label{fig:teaser_b}
  \end{subfigure}
  \caption{\textbf{The paradox of visual tool-use.} \textbf{(a)} Our experiments show that, despite far more tokens, tool-use policies yield marginal or no improvements over direct inference. \textbf{(b)} A qualitative failure on a V$^{\!*}$ instance (by Mini-o3): the direct prediction is correct, but tool-augmented inference repeatedly crops irrelevant regions and leads to no valid answer or a wrong answer.}
  \label{fig:teaser_paradox}
\end{figure}

While thinking-with-images is intuitive, in practice it often yields only marginal or even negative accuracy improvements over direct inference at substantially higher token cost (Fig.~\ref{fig:teaser_paradox}(a); see also~\citealt{wei2026zooming}), and tool-augmented rollouts sometimes fail on questions a direct prediction would have answered correctly, by repeatedly cropping into irrelevant regions (Fig.~\ref{fig:teaser_paradox}(b)). These observations raise a key question: \textbf{how does visual evidence returned by tools actually affect the model's final prediction?}

\begin{figure*}[t]
  \centering
  \includegraphics[width=\textwidth]{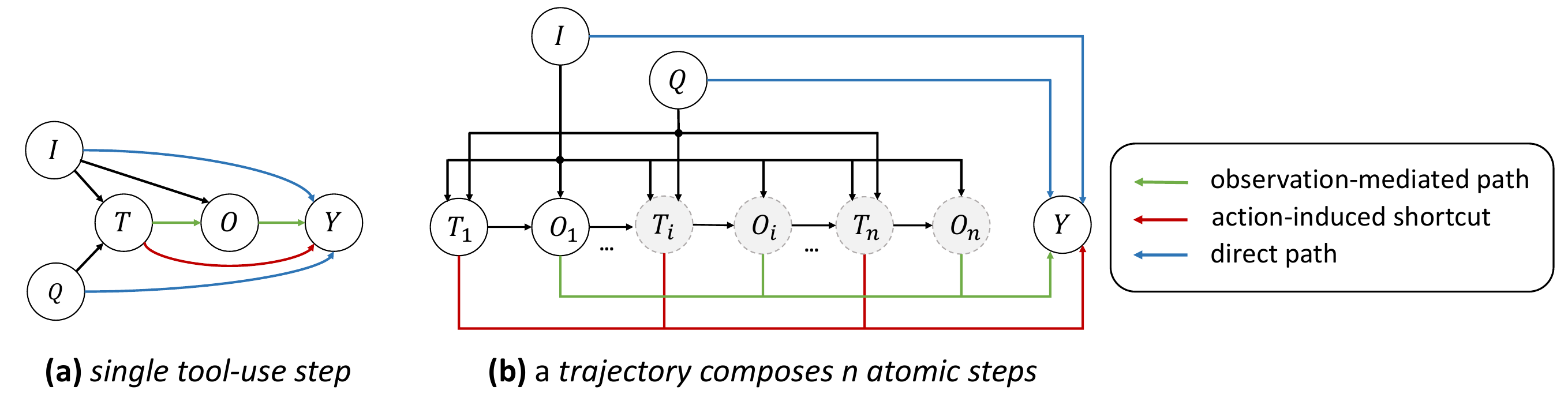}
  \caption{\textbf{Causal graph for visual tool-use.} \textbf{(a)} The single-step causal graph for a single tool-use step over $(I, Q, T, O, Y)$, highlighting the direct path $(I,Q)\!\to\!Y$ (blue), the observation-mediated path $T\!\to\!O\!\to\!Y$ (green) and the action-induced shortcut $T\!\to\!Y$ (red).  \textbf{(b)} A full trajectory composes $n$ such atomic graphs. }
  \label{fig:causal_framework}
\end{figure*}

To study this question, we formulate visual tool-use as a causal graph (Fig.~\ref
{fig:causal_framework}) over the input image $I$, query $Q$, the policy-emitted tool 
actions $T_i$ and returned observations $O_i$ at each step $i$, and the final answer 
$Y$. Beyond the direct path $(I,Q)\!\to\!Y$, a tool-use trajectory distinguishes two types of paths from $(I,Q)$ to $Y$: (i) \textbf{observation-mediated paths} that carry \emph
{genuine grounding}, where the visual content of some $O_i$ propagates to $Y$; (ii) 
\emph{undesired} \textbf{action-induced shortcuts} that reach $Y$ purely through $T$, 
so the textual action alone influences the answer.

We then design three interventions at progressively finer granularity. A \emph{policy-level} intervention (\S\ref{sec:policy_level_intervention}) toggles tool-use against direct inference. A \emph{trajectory-level} intervention (\S\ref{sec:trajectory_level_intervention}) corrupts every returned observation $O_i$ at runtime and measures the resulting change in the observation-feedback process. A \emph{step-level} intervention (\S\ref{sec:step_level_intervention}) replaces one observation $O_i$ with a counterfactual crop and probes whether its visual content changes the model's immediate preference over the answer options.

Applying this formulation to six representative thinking-with-images models across five fine-grained perception benchmarks, we uncover a structural decoupling between the tool-use policy and the actual utility of the observations it elicits. We term it \textbf{\emph{policy miscalibration}} and identify two failure modes: in \emph{Mode~1: Calling Without Looking (CWL)} the visual content of the observation has no causal effect on the final answer; in \emph{Mode~2: Looking Without Planning (LWP)} the visual content does shape the answer, but the tool-calling strategy is poorly planned (continuing after the answer is already correct, or repeatedly cropping into uninformative regions).

The two failure modes of policy miscalibration (\emph{CWL} and \emph{LWP}), together with rollouts that emit no tool calls (\emph{No-call}) and rollouts that use tools effectively (\emph{Calibrated}), form the basis of a trajectory-level \textbf{diagnostic} that sorts every rollout into one of these four groups.
We decompose the policy-level accuracy improvement into per-group contributions and
find that the accuracy improvements are mainly driven by the \emph{Calibrated} subset. We call this \textbf{\emph{the illusion of visual tool-use}}: benchmark improvements read as effective tool-use, but are in fact concentrated in one calibrated minority of rollouts, while the majority of tool calls is either decoupled from the answer or scheduled incoherently.

Finally, we propose a hypothesis that outcome-only reinforcement learning causes the failure and discuss the operational uses of the diagnostic.

\paragraph{Contributions.}
(i)~We formulate visual tool-use as a causal graph and design a three-level intervention protocol (policy, trajectory, step) that disentangles \emph{observation-mediated paths} from \emph{action-induced shortcuts} (\S\ref{sec:causal_framework}).
(ii)~We identify \emph{policy miscalibration} as the central bottleneck of current thinking-with-images models, with two failure modes (\emph{CWL} and \emph{LWP}), build a trajectory-level diagnostic that sorts every rollout into one of four groups, and decompose the policy-level accuracy improvement into per-group contributions (\S\ref{sec:diagnosing_policy_miscalibration}).
(iii)~We characterize \emph{the illusion of visual tool-use}, propose an RL-trap hypothesis attributing policy miscalibration to outcome-only reinforcement learning, and discuss operational uses of the diagnostic (\S\ref{sec:discussion}).

\section{Related Work}

\paragraph{Thinking with images and visual tool-use.}
A growing line of work equips MLLMs with active visual operations such as crop-and-zoom~\citep{openai2025thinkingwithimages,su2025thinkingwithimages,zhang2025causight,zheng2025deepeyes,wang2025pixel,lai2025mini-o3,bai2025qwen3vl,zhang2026thyme}, with gains commonly attributed to higher-resolution evidence. Recent studies challenge this claim: \citet{wei2026zooming} report a marginal-improvement paradox; \citet{ma2026does} attribute most improvement to intrinsic learning rather than tool mastery; and \citet{hou2025codev,xu2025defacto} expose unfaithful tool-use. These analyses stay at the policy or behavioral level; we instead push the diagnosis to the step level via a causal view and quantitatively attribute where the improvements come from.

\citet{wang2025virl} describe an ``illusion of thinking with images'': visual actions may look like rationales while remaining weakly grounded, and propose process-aware training to improve them. Our ``illusion of visual tool-use'' is instead a diagnostic claim: aggregate gains can create the impression that visual tool-use is broadly effective, even though returned observations do not causally mediate the answer on many tool-using trajectories. Process supervision can improve call quality, but does not decide whether a call is needed; ViRL still averages about one visual action per example, while many cases are already solvable by direct inference.

\paragraph{Causal analysis and faithfulness of LLM reasoning.}
A parallel line probes whether intermediate artifacts of LLM reasoning are causally tied to the final answer rather than post-hoc rationalization~\citep{turpin2023faithful,lanham2023measuring,karimi2023relationship,chen2024quantifying,chen2024cello,bentham2024chain,chen2025imitation,han2025beyond}. Our framework adapts Pearl's do-intervention~\citep{pearl2009causality} to multimodal tool-use traces, and VEG instantiates the natural indirect effect~\citep{pearl2001direct,robins1992identifiability,avin2005identifiability} via observation interventions.

\section{Causal Framework for Visual Tool-Use}
\label{sec:causal_framework}

\subsection{Preliminaries}
\label{sec:preliminaries}

\paragraph{Visual tool-use.}

We study \emph{visual tool-use}, an instance of the thinking-with-images paradigm~\citep{openai2025thinkingwithimages,su2025thinkingwithimages} in which an MLLM augments its reasoning with calls to a visual engine. Given image $I$ and query $Q$, at step $i$ the policy emits a tool action $T_i \sim \pi(\cdot \mid I, Q, T_{<i}, O_{<i})$ which decides the region to \textsc{crop-and-zoom}~\citep{zheng2025deepeyes,wang2025pixel,lai2025mini-o3}. A visual engine $E_{\text{tool}}$ then executes the action and returns the cropped sub-image $O_i $. After $n$ steps the MLLM commits a final answer $Y$ conditioned on $(I,Q,T_{1:n},O_{1:n})$.
\paragraph{Causal inference primitives.}
We work in Pearl's structural causal model (SCM)~\citep{pearl2009causality}. For a variable $X$, the intervention $do(X{=}x)$ overrides $X$ to $x$ and cuts its incoming edges; the \textit{Average Treatment Effect (ATE)} of switching $X$ from $x$ to $x'$ on outcome $Y$ is $\mathbb{E}[Y\mid do(X{=}x')]-\mathbb{E}[Y\mid do(X{=}x)]$. When $X$ influences $Y$ both directly and through a mediator $M$, the \textit{Natural Indirect Effect (NIE)}~\citep{pearl2001direct,robins1992identifiability,avin2005identifiability} isolates the $X\!\to\!M\!\to\!Y$ component as $\mathbb{E}[Y\mid do(X{=}x,M{=}M(x'))]-\mathbb{E}[Y\mid do(X{=}x,M{=}M(x))]$, where $M(x)$ is the value $M$ takes under $do(X{=}x)$. Our step-level estimand \textit{Visual Evidence Gain} (\S\ref{sec:step_level_intervention}) measures this NIE along $T\!\to\!O\!\to\!Y$ through a counterfactual intervention.

\subsection{Causal Formulation of Visual Tool-Use}
\label{sec:causal_formulation}
We formulate a single rollout as an SCM over $(I,Q,T_{1:n},O_{1:n},Y)$ (Fig.~\ref{fig:causal_framework}), with mechanisms
\begin{equation*}
\begin{aligned}
T_i &\leftarrow f_\pi(I,Q,T_{<i},O_{<i}),\\
O_i &\leftarrow E_{\text{tool}}(I,T_i),\\
Y   &\leftarrow f_\pi(I,Q,T_{1:n},O_{1:n}),
\end{aligned}
\end{equation*}
where the same $f_\pi$ (induced by the MLLM policy $\pi$) produces both tool actions and the final answer.

As shown in Fig.~\ref{fig:causal_framework}, there are three types of paths from input $(I,Q)$ to $Y$: (i) the baseline direct path $(I,Q)\!\to\!Y$ without tools; (ii) the intended \emph{observation-mediated path} through any $O_i$, where $T_i$ selects a region and the resulting crop $O_i$ informs the answer; (iii) the undesired \emph{action-induced shortcut} purely through the action trace $T_{1:n}$, along which the mere presence of a call shifts the answer regardless of $O$ (e.g.,\ a zoom call boosts confidence in a prior hypothesis).

\subsection{Three-Level Causal Interventions}
\label{sec:three_level_causal_interventions}

\begin{figure*}[t]
  \centering
  \includegraphics[width=\textwidth]{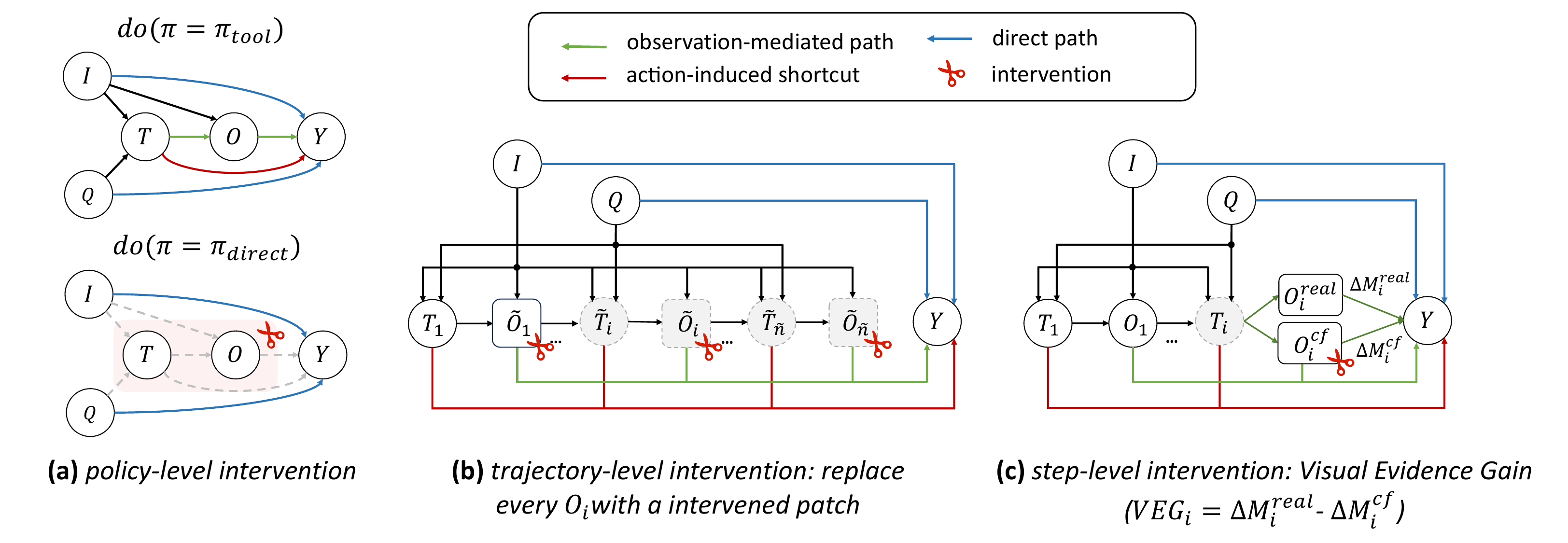}
  \caption{\textbf{Three-level causal intervention.} \textbf{(a)} \emph{Policy-level intervention} (\S\ref{sec:policy_level_intervention}) toggles the entire $T\!\to\!O\!\to\!Y$ subgraph. \textbf{(b)} \emph{Trajectory-level intervention} (\S\ref{sec:trajectory_level_intervention}) replaces every real observation with a corrupted one $\tilde O_i$ during inference. The policy reacts to corrupted feedback and may produce a different action trace $\tilde T_{1:\tilde n}$. \textbf{(c)} \emph{Step-level intervention} (\S\ref{sec:step_level_intervention}) replaces the real observation $O_i^{\text{real}}$ with a counterfactual one $O_i^{\text{cf}}$; $\text{VEG}_i = \Delta M^{\text{real}}_i - \Delta M^{\text{cf}}_i$ is a local counterfactual-based estimate of the \emph{Natural Indirect Effect} along $T_i\!\to\!O_i\!\to\!Y$, capturing the contribution of the returned visual evidence.}
  \label{fig:causal_intervention}
\end{figure*}

Building on the causal graph above, we intervene at three levels (policy, trajectory, and step) to quantify the causal contribution of visual tool-use at progressively finer granularity.

\subsubsection{Policy-Level Intervention: Tool-Use vs.\ Direct Inference}
\label{sec:policy_level_intervention}

We begin by asking: \textit{can visual tool-use improve model accuracy compared with direct inference?}
\paragraph{Estimand and setup.}
We compare the tool-enabled policy $\pi_{\text{tool}}$ with a direct policy $\pi_{\text{direct}}$ via $do(\pi{=}\pi_{\text{tool}})$ vs.\ $do(\pi{=}\pi_{\text{direct}})$ (Fig.~\ref{fig:causal_intervention}(a)), where $\pi_{\text{direct}}$ uses a tool-free system prompt and otherwise shares weights and decoding with $\pi_{\text{tool}}$. The policy-level ATE on accuracy is
\begin{equation}
\label{eq:policy_ate}
\begin{aligned}
\mathrm{ATE}_{\text{policy}}
&= \mathbb{E}_{do(\pi{=}\pi_{\text{tool}})}
   \!\left[\mathbf{1}[ Y{=}Y^\star]\right] \\
&\quad - \mathbb{E}_{do(\pi{=}\pi_{\text{direct}})}
   \!\left[\mathbf{1}[ Y{=}Y^\star]\right],
\end{aligned}
\end{equation}
with $Y^\star$ the ground-truth answer.
We evaluate six open visual tool-use models covering both implementation families: interface-based DeepEyes~\citep{zheng2025deepeyes}, Pixel Reasoner~\citep{wang2025pixel}, Mini-o3~\citep{lai2025mini-o3}, and Qwen3-VL-4B/8B~\citep{bai2025qwen3vl}, and code-based Thyme~\citep{zhang2026thyme}, on V$^{\!*}$~\citep{wu2024v}, HR-Bench (4K/8K)~\citep{wang2025hrbench}, VisualProbe~\citep{lai2025mini-o3}, and MME-RealWorld-Lite~\citep{zhang2025mmerealworld}. See App.~\ref{app:experimental_setup} for details.

\begin{table*}[t]
\caption{\textbf{Policy-level intervention: tool-use vs.\ direct inference.}
\emph{Acc} reports accuracy (\%) under $\pi_{\text{tool}}$\,/\,$\pi_{\text{direct}}$;
$\Delta$ is the policy-level ATE in percentage points (pp) (Eq.~\ref{eq:policy_ate}). VisualProbe combines Easy/Medium/Hard levels.}
\label{tab:policy_ate}
\centering
\small
\setlength{\tabcolsep}{3.5pt}
\begin{tabular}{llccccc}
\toprule
\multicolumn{2}{l}{Model} & V$^{\!*}$ & HR-Bench-4K & HR-Bench-8K & VisualProbe & MME-RealWorld-Lite \\
\midrule
\multicolumn{7}{l}{\emph{Interface-based}} \\
DeepEyes & Acc & 83.3/83.3 & 75.0/76.4 & 72.0/72.6 & 42.9/42.9 & 54.7/54.9 \\
         & $\Delta$ & $0.0$ & $-1.4$ & $-0.6$ & $+0.0$ & $-0.2$ \\
\addlinespace[2pt]
Pixel Reasoner & Acc & 84.8/81.2 & 74.0/74.4 & 68.2/64.0 & 37.3/29.7 & 64.4/54.4 \\
               & $\Delta$ & $+3.6$ & $-0.4$ & $+4.2$ & $+7.6$ & $+10.0$ \\
\addlinespace[2pt]
Mini-o3 & Acc & 87.8/82.3 & 76.8/72.0 & 73.2/67.4 & 55.0/33.7 & 65.5/53.6 \\
        & $\Delta$ & $+5.5$ & $+4.8$ & $+5.8$ & $+21.3$ & $+11.9$ \\
\addlinespace[2pt]
Qwen3-VL-8B & Acc & 91.1/84.2 & 83.9/78.9 & 79.5/75.7 & 48.6/36.9 & 51.0/54.7 \\
            & $\Delta$ & $+6.9$ & $+5.0$ & $+3.8$ & $+11.7$ & $-3.7$ \\
\addlinespace[2pt]
Qwen3-VL-4B & Acc & 86.4/82.0 & 80.1/80.1 & 75.3/76.9 & 45.7/38.9 & 49.0/50.3 \\
            & $\Delta$ & $+4.4$ & $0.0$ & $-1.6$ & $+6.8$ & $-1.3$ \\
\midrule
\multicolumn{7}{l}{\emph{Code-based}} \\
Thyme & Acc & 83.2/80.2 & 78.3/74.7 & 73.0/68.2 & 46.2/43.7 & 55.2/48.9 \\
      & $\Delta$ & $+3.0$ & $+3.6$ & $+4.8$ & $+2.6$ & $+6.3$ \\
\bottomrule
\end{tabular}
\end{table*}

\paragraph{Results.}
Table~\ref{tab:policy_ate} shows that the policy-level ATE varies sharply across models: essentially null on DeepEyes, modest on Pixel Reasoner / Qwen3-VL-4B / Thyme, and largest on Mini-o3 and Qwen3-VL-8B (e.g., $+21.3$ pp on VisualProbe). The improvements are thus limited and uneven; moreover, the policy-level ATE conflates observation-mediated paths with action-induced shortcuts, leaving open whether the improvement is genuinely carried by visual evidence, motivating the next two interventions.

\subsubsection{Trajectory-Level Intervention: Dynamic Observation Corruption}
\label{sec:trajectory_level_intervention}

\begin{table*}[t]
\caption{\textbf{Trajectory-level intervention under \textsc{random-crop}.}
\emph{Acc} reports accuracy (\%) under
$\pi_{\text{tool}}$\,/\,the dynamic intervention
$do(O_i{=}\tilde O_i\,\forall i)$; $\Delta$ is
the trajectory-level ATE on
accuracy in pp (Eq.~\ref{eq:traj_delta}).
The rightmost column reports \emph{Hit-MaxT}, the fraction of rollouts that hit the tool-call limit.}
\label{tab:traj_results}
\centering
\small
\setlength{\tabcolsep}{3pt}
\begin{tabular}{llccccc c}
\toprule
\multicolumn{2}{l}{Model} & V$^{\!*}$ & HR-Bench-4K & HR-Bench-8K & VisualProbe & MME-RealWorld-Lite & Hit-MaxT (V$^{\!*}$) \\
\midrule
\multicolumn{8}{l}{\emph{Interface-based}} \\
DeepEyes        & Acc      & 83.3/83.8 & 75.0/74.1 & 72.0/70.4 & 42.9/44.2 & 54.7/54.6 & \multirow{2}{*}{$0\%$} \\
                & $\Delta$ & $+0.5$    & $-0.9$    & $-1.6$    & $+1.3$    & $-0.1$ & \\
\addlinespace[2pt]
Pixel Reasoner  & Acc      & 84.8/82.7 & 74.0/70.8 & 68.2/66.1 & 37.3/33.7 & 64.4/49.3 & \multirow{2}{*}{$0\%$} \\
                & $\Delta$ & $-2.1$    & $-3.2$    & $-2.1$    & $-3.6$    & $-15.1$ & \\
\addlinespace[2pt]
Mini-o3         & Acc      & 87.8/23.6 & 76.8/20.4 & 73.2/15.8 & 55.0/2.4  & 65.5/15.1 & \multirow{2}{*}{$84.8\%$} \\
                & $\Delta$ & $-64.2$   & $-56.4$   & $-57.4$   & $-52.6$   & $-50.4$ & \\
\addlinespace[2pt]
Qwen3-VL-8B     & Acc      & 91.1/30.4 & 83.9/46.4 & 79.5/41.0 & 48.6/11.2 & 51.0/29.7 & \multirow{2}{*}{$59.2\%$} \\
                & $\Delta$ & $-60.7$   & $-37.5$   & $-38.5$   & $-37.4$   & $-21.3$ & \\
\addlinespace[2pt]
Qwen3-VL-4B     & Acc      & 86.4/38.2 & 80.1/59.3 & 75.3/46.6 & 45.7/8.8  & 49.0/34.3 & \multirow{2}{*}{$50.3\%$} \\
                & $\Delta$ & $-48.2$   & $-20.8$   & $-28.7$   & $-36.9$   & $-14.7$ & \\
\midrule
\multicolumn{8}{l}{\emph{Code-based}} \\
Thyme           & Acc      & 83.2/82.2 & 78.3/77.9 & 73.0/74.1 & 46.2/45.6 & 55.2/56.2 & \multirow{2}{*}{$0\%$} \\
                & $\Delta$ & $-1.0$    & $-0.4$    & $+1.1$    & $-0.6$    & $+1.0$ & \\
\bottomrule
\end{tabular}
\end{table*}

Next, as a coarse trajectory-level test, we ask: \textit{what is the total effect of corrupting returned observations during rollout?}

\paragraph{Estimand and setup.} We perform a \emph{dynamic observation intervention} (Fig.~\ref{fig:causal_intervention}(b)) by replacing each requested tool output with a corrupted observation $\tilde O_i$ during rollout. Since subsequent actions condition on earlier observations, the intervention may also alter later actions and stopping decisions. Therefore the trajectory-level ATE measures the total effect of corruption, including both evidence removal and induced policy reactions such as repair loops and budget exhaustion.

\begin{equation}
\label{eq:traj_delta}
\begin{aligned}
\mathrm{ATE}_{\text{traj}}
&= \mathbb{E}\!\left[\mathbf{1}[Y{=}Y^\star]\,\big|\,
   do(O_i{=}\tilde O_i\,\forall i)\right] \\
&\quad - \mathbb{E}\!\left[\mathbf{1}[ Y{=}Y^\star]\right].
\end{aligned}
\end{equation}

We use the same models and benchmarks as \S\ref{sec:policy_level_intervention}. Our primary corruption, \textsc{random-crop}, returns a same-shape crop sampled from another location in $I$. It preserves the modality, source image, and crop-like format, reducing distribution shift relative to \textsc{noise} and \textsc{blank}. The \textsc{noise} and \textsc{blank} ablations are reported in App.~\ref{app:corrupt-vstar}. We also report \emph{Hit-MaxT}, the fraction of intervened trajectories that reach the tool-call limit.

\paragraph{Results.} Table~\ref{tab:traj_results} reports the effects of intervention. DeepEyes and Thyme show near-zero net effects across benchmarks. Pixel Reasoner shows modest effects in most settings, with a larger drop on MME-RealWorld-Lite. Mini-o3 and both Qwen3-VL models show large negative effects across benchmarks. Their reasoning parts often identify that the returned observation is inconsistent with the request, then issue further calls that frequently exhaust the budget.

To control for the portion of the drop caused solely by answer truncation, we add a force-answer experiment (App.~\ref{app:force-answer}). It preserves the corrupted-observation history but requires a final answer when the call limit is reached, removing failures caused only by a missing answer. The experiment confirms that truncation accounts for part of the drop. Nevertheless, accuracy remains substantially below clean tool use, showing that truncation is not the sole cause. This residual effect is consistent with the loss of useful visual evidence, although other policy reactions may also contribute.

\subsubsection{Step-Level Intervention: Visual Evidence Gain}
\label{sec:step_level_intervention}

\begin{table*}[t]
\caption{\textbf{Step-level intervention on V$^{\!*}$.}
\emph{Marginal Gain (MG)} is reported with the original observation ($\Delta M^{\text{real}}$) and with the counterfactual observation ($\Delta M^{\text{cf}}$).
\emph{Visual Evidence Gain (VEG)} isolates the contribution of the visual content; we report it pooled over \textbf{A}ll calls and split by whether the trajectory is ultimately \textbf{C}orrect or \textbf{I}ncorrect.
\emph{Distribution} shows the fraction of calls that are near-zero ($|\mathrm{VEG}|\!<\!0.01$) or substantial ($|\mathrm{VEG}|\!>\!0.1$).
\emph{By saturation} conditions VEG on pre-call confidence: \emph{Sat-rate} is the fraction of calls whose pre-call probability gap $g_{i-1}$ already exceeds $0.95$, and \emph{Sat}/\emph{Non-sat} are mean VEG conditional on (non-)saturation.
\emph{mean $V^{\max}$} is each trajectory's best per-call VEG, averaged across trajectories, capturing how much the most useful call in a trajectory contributes on average.}
\label{tab:veg_results}
\centering
\small
\setlength{\tabcolsep}{2.5pt}
\begin{tabular}{l cc ccc cc ccc c}
\toprule
& \multicolumn{2}{c}{MG (mean)} & \multicolumn{3}{c}{VEG all-calls (mean)} & \multicolumn{2}{c}{Distribution} & \multicolumn{3}{c}{By saturation} & \\
\cmidrule(lr){2-3}\cmidrule(lr){4-6}\cmidrule(lr){7-8}\cmidrule(lr){9-11}
Model & $\Delta M^{\text{real}}$ & $\Delta M^{\text{cf}}$ & A & C & I & $|\mathrm{VEG}|\!<\!0.01$ & $|\mathrm{VEG}|\!>\!0.1$ & Sat-rate & Sat & Non-sat & mean $V^{\max}$ \\
\midrule
DeepEyes      & $0.412$ & $0.359$ & $0.053$ & $+0.07$ & $-0.01$ & $81\%$ & $13\%$ & $37\%$ & $-0.02$ & $0.10$ & $+0.05$ \\
\addlinespace[2pt]
Mini-o3       & $0.113$ & $0.051$ & $0.063$ & $+0.09$ & $-0.01$ & $29\%$ & $34\%$ & $21\%$ & $\sim\!0$ & $0.08$ & $+0.19$ \\
\addlinespace[2pt]
Qwen3-VL-8B   & $0.200$ & $0.001$ & $0.198$ & $+0.28$ & $-0.20$ & $56\%$ & $29\%$ & $64\%$ & $\sim\!0$ & $0.55$ & $+0.30$ \\
\bottomrule
\end{tabular}
\end{table*}

\begin{figure*}[t]
\centering
\includegraphics[width=0.7\textwidth]{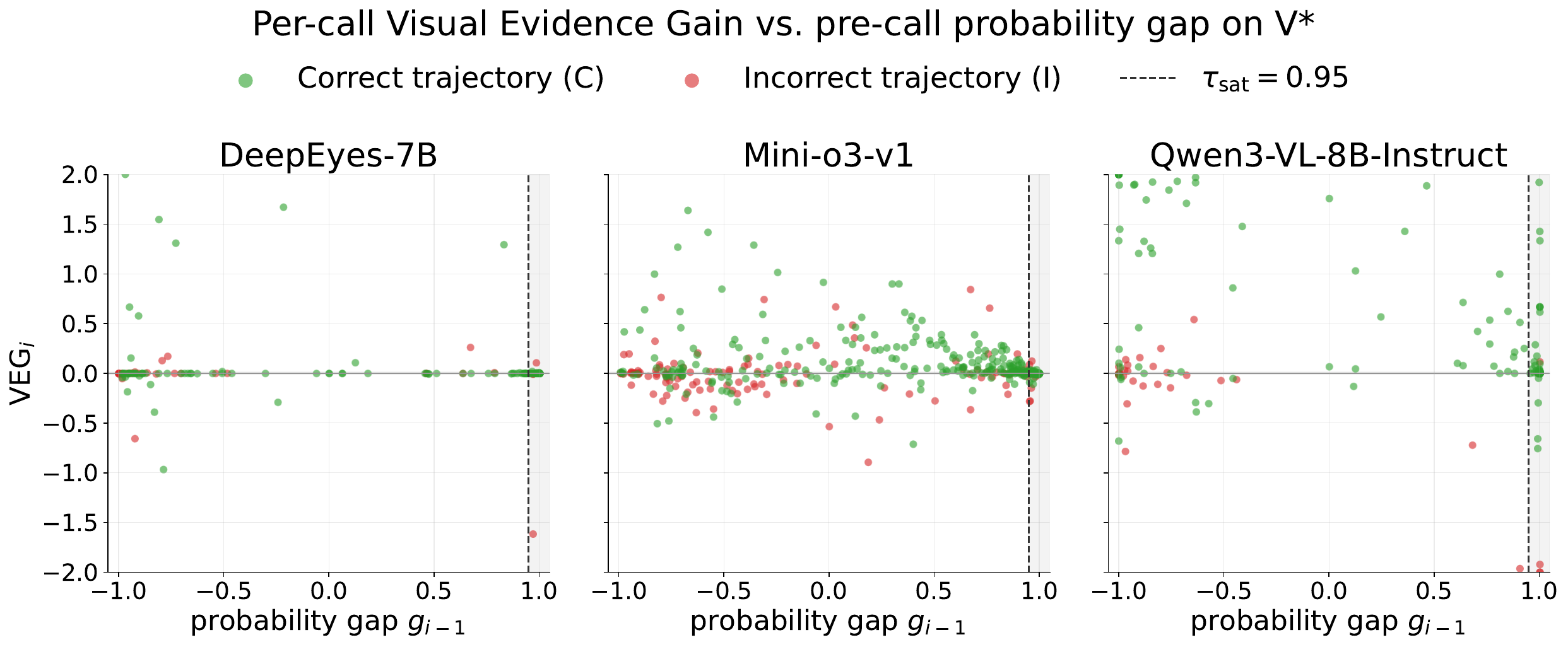}
\caption{\textbf{Per-call scatter of probability gap before call $g_{i-1}$ vs.\ $\mathrm{VEG}_i$ on V$^{\!*}$.} Each point is one tool call; color indicates the trajectory's final correctness (\textbf{C}/\textbf{I}); dashed line marks $\tau_{\text{sat}}\!=\!0.95$.
}
\label{fig:veg_scatter}
\end{figure*}

Finally, we ask: \textit{how much does each observation $O_i$ contribute to the final answer?}
\paragraph{Estimand.} \textbf{1) Visual Evidence Gain}
We probe the observation-mediated path $T_i\!\to\!O_i\!\to\!Y$ with a fixed next-token readout rather than an additional free-form answer-probing rollout.
At each checkpoint $i$, we take the generated prefix through the current thinking block, append the literal string \texttt{<answer>}, and read the model's next-token logits $\{z_y\}$.
Let $V$ be the candidate label set (usually $\{\texttt{A},\texttt{B},\texttt{C},\texttt{D}\}$).
The option-restricted softmax is $\tilde P(y\mid C)=\exp(z_y)/\sum_{k\in V}\exp(z_k)$.
With $Y^\star$ the correct option and $y_{\mathrm{err}}=\arg\max_{k\neq Y^\star}\tilde P(k\mid C)$ the strongest distractor, the \emph{probability gap} is
\begin{equation}
\label{eq:prob_gap}
g(C)=\tilde P(Y^\star\mid C)-\tilde P(y_{\mathrm{err}}\mid C)\in[-1,1].
\end{equation}
We apply this readout in three contexts at step $i$: before the call; after the tool action $T_i$ and the real observation $O_i^{\text{real}}$; and after the same $T_i$ and a counterfactual observation $O_i^{\text{cf}}$.
We write $g_{i-1}$ for the gap before the call and $g_i^{\text{real}}$ and $g_i^{\text{cf}}$ for the gaps after the real and counterfactual observations.
The per-step \emph{Marginal Gain} is $\Delta M_i^{\text{real}} = g_i^{\text{real}} - g_{i-1}$ and $\Delta M_i^{\text{cf}} = g_i^{\text{cf}} - g_{i-1}$.
The counterfactual $O_i^{\text{cf}}$ is a same-shape crop sampled from a different location in $I$ (Fig.~\ref{fig:causal_intervention}(c)).
The \emph{Visual Evidence Gain} is
\begin{equation}
\label{eq:veg}
\mathrm{VEG}_i \;=\; \Delta M_i^{\text{real}} - \Delta M_i^{\text{cf}}.
\end{equation}
$\mathrm{VEG}_i$ is a counterfactual step-level probing estimate of observation-mediated influence.
Because the pre-call prefix, $T_i$, and \texttt{<answer>} marker are identical under the real and counterfactual observations, the action-induced shortcut $T_i\!\to\!Y$ cancels and $\mathrm{VEG}_i$ measures the local contribution of the returned visual content.

\noindent \textbf{2) Saturation Rate}
However, $\mathrm{VEG}_i$ can be near zero not only because the visual content is uninformative, but also because the performance gap $g_{i-1}$ in previous steps is already high, leaving little room for step $i$ to improve.
We call this situation \emph{saturated}: $g_{i-1}>\tau_{\text{sat}}$, where $\tau_{\text{sat}}$ is the saturation threshold. 
Then, we report VEG (a)~average pooled over all calls, (b)~conditional on saturation, and (c)~split by trajectory correctness (\textbf{C}/\textbf{I}). Moreover, we report the fraction of \emph{near-zero} ($|\mathrm{VEG}|\!<\!0.01$) and \emph{substantial} ($|\mathrm{VEG}|\!>\!0.1$) calls.

\noindent \textbf{3) Post-saturation Over-Extension Rate}: We further analyze how often the policy continues calling after it is saturated with the \emph{Post-saturation Over-Extension Rate}.
\begin{equation}
\label{eq:poer}
\mathrm{POER}=\Pr\!\big(\,g_{k-1}\!>\!\tau_{\text{sat}}\,\big|\,k\!\geq\!2,\,g_0\!\le\!\tau_{\text{sat}}\big).
\end{equation}
Any call made after reaching the saturation threshold is termed an \emph{over-extension} (more details in App.~\ref{app:veg_full}).

\paragraph{Setup.}
For multiple-choice questions, $g_i$ is read from option-restricted softmax.
We set $\tau_{\text{sat}}\!=\!0.95$. The main text reports three interface-based models (DeepEyes, Mini-o3, Qwen3-VL-8B) chosen to span the policy behaviors surfaced by \S\ref{sec:policy_level_intervention}--\S\ref{sec:trajectory_level_intervention}. The open-ended analogue that replaces this readout with a length-normalized gold log-likelihood is in App.~\ref{app:veg_openended}.

\paragraph{Results.}
\noindent \textbf{1) Visual evidence contribution}
The non-zero $\Delta M^{\text{cf}}$ column in Table~\ref{tab:veg_results} shows that even an irrelevant crop pushes the answer toward the ground truth purely due to the call itself; only the residual $\Delta M^{\text{real}}\!-\!\Delta M^{\text{cf}}$ recovers the genuine visual contribution, motivating VEG. The per-call decomposition (Table~\ref{tab:veg_results}; scatter in Fig.~\ref{fig:veg_scatter}) reveals three distinct model behaviors:
\begin{enumerate}[leftmargin=*,itemsep=2pt,topsep=2pt]
   \item \textbf{Structurally inactive} (DeepEyes): across all columns (all, correct, incorrect, saturated, non-saturated), VEG is around zero, and even the best call in a trajectory does nothing, indicating that the visual evidence contributes little.
   \item \textbf{Information-driven but diluted} (Mini-o3):
   A positive per-trajectory peak $V^{\max}$ indicates Mini-o3 \emph{can} produce tool calls with effective visual evidence. However, the pooled mean score is low, reflecting that most calls are still ineffective. The mean is also nearly the same with or without saturation, so usefulness barely depends on pre-call confidence. 
   \item \textbf{Information-driven, sharply concentrated} (Qwen3-VL-8B): The usefulness of calls depends on the saturation. Saturated calls carry no signal, non-saturated calls carry large magnitude; the per-trajectory peak is strongly positive (negative) on correct (incorrect) trajectories. 
\end{enumerate}

\begin{figure}[t]
\centering
\includegraphics[width=\columnwidth]{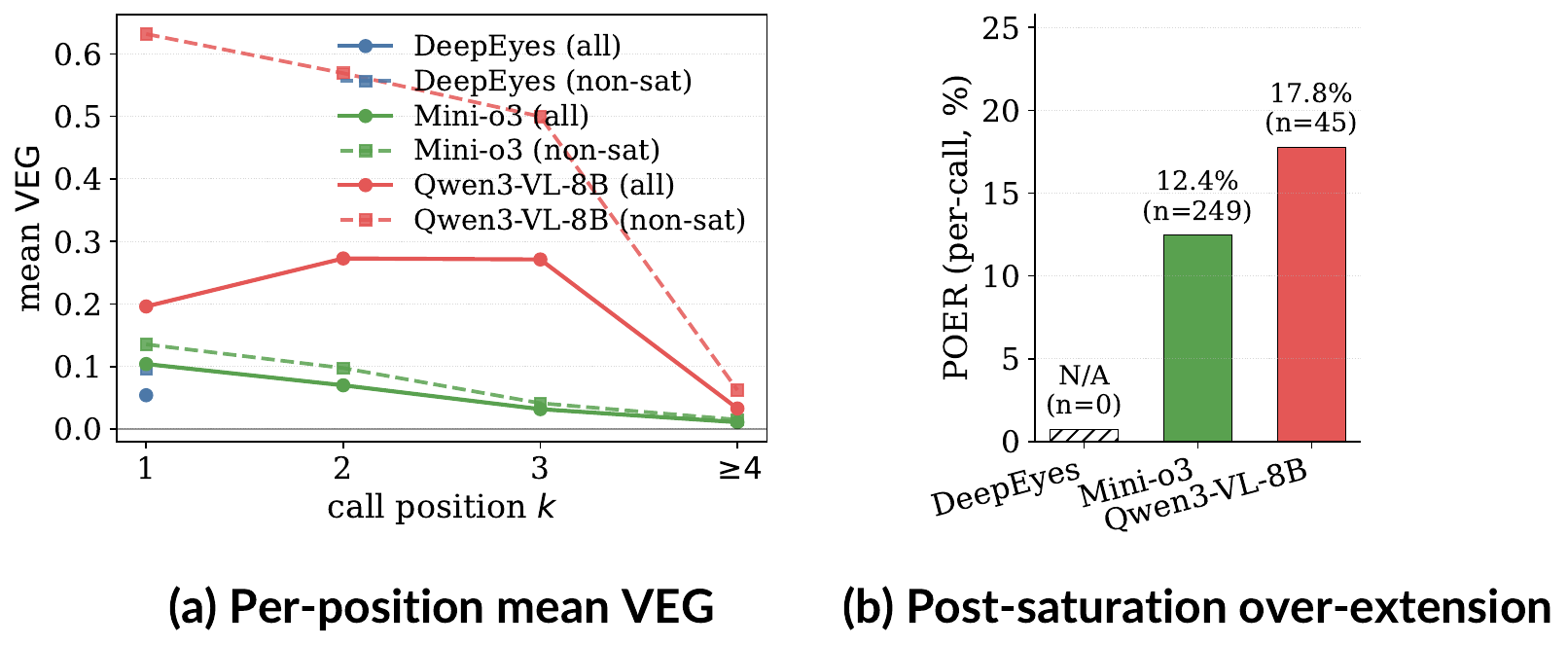}
\caption{\textbf{Within-trajectory dynamics on the V$^{\!*}$ benchmark}. (a)~Mean VEG by call position $k\!\in\!\{1,2,3,\geq\!4\}$; Solid curves indicate mean on all calls and dashed curves indicate mean on non-saturated subset only. (b)~Per-position Over-Extension Rate (POER, Eq.~\ref{eq:poer}, per-call). Full per-position breakdown in App.~\ref{app:per_pos_detail}.}
\label{fig:per_pos_dynamics}
\end{figure}

\noindent \textbf{2) Within-trajectory Dynamics} Figure~\ref{fig:per_pos_dynamics}(a) plots per-step (per-position) mean VEG; the near-alignment of the dashed (non-saturated) and solid (all-call) curves indicates that the VEG decay is not caused by later calls being saturated. Three distinct model behaviors emerge: \textbf{Mini-o3} shows \emph{uniform per-call decay}: the mean VEG decreases steadily with each position $k$, and over-extension occurs evenly across positions; \textbf{Qwen3-VL-8B} shows \emph{sustained value with early over-extension}: non-saturated VEG remains substantially higher than Mini-o3's for $k\!\leq\!3$, while every over-extension call lands early at $k\!=\!2,3$; \textbf{DeepEyes}: nearly all trajectories contain only a single tool call, making POER undefined as it requires $k\geq 2$.

\section{Diagnosing Policy Miscalibration}
\label{sec:diagnosing_policy_miscalibration}

The interventions above reveal a consistent finding: \textbf{the policy's decisions over \emph{when} to call, \emph{when} to stop, and \emph{when} to commit are decoupled from the actual causal utility of the resulting observations}. Tools are invoked when their evidence cannot help, and tool calls continue even after the model already has sufficient confidence to answer. We call this \emph{\textbf{policy miscalibration}} and characterize two failure modes.

\subsection{Two Failure Modes}
\label{sec:two_failure_modes}
\paragraph{Mode~1: Calling Without Looking (CWL).}
A tool call is emitted, yet the visual content of its observation does not contribute to the answer, manifesting two sub-cases: \emph{(a)~Saturated prior}: the policy calls a tool when $g_{i-1}\!>\!\tau_{\text{sat}}$ already, so any contribution is mechanically bounded near zero (dominant for Qwen3-VL-8B). \emph{(b)~Structurally inactive call}: no call along the rollout carries visual evidence beyond the action-induced shortcut (exemplified by DeepEyes). In both cases the per-step gain is essentially carried by $T\!\to\!Y$; tool-use behaves as a syntactic ritual rather than a perceptual act.

\paragraph{Mode~2: Looking Without Planning (LWP).}
Per-call VEG is non-zero and the observation-mediated path is active, but planning is incoherent. Two situations: \emph{(a)~Post-saturation extension}: the policy keeps calling after $g_{i-1}$ crosses $\tau_{\text{sat}}$ (Mini-o3's per-position VEG decay with rising harmful-rate; Qwen3-VL-8B's early saturation with continued calls). \emph{(b)~Budget exhaustion}: the policy never stops on its own and exhausts the tool-call budget.

\subsection{Trajectory-Level Diagnostic}
\label{sec:per_trajectory_diagnostic}
We construct a diagnostic classifier with four groups: trajectories with the two failure modes of policy miscalibration (\emph{CWL} and \emph{LWP}), trajectories without any tool calls (defined as \emph{No-call}), and trajectories that use tools effectively (defined as \emph{Calibrated}).
For a trajectory $\tau$ we compute five features on the vanilla rollout: the number of tool calls $n$; the pre-first-call probability gap $g_0$ ($+\infty$ if $n\!=\!0$); the peak per-call VEG $V^{\max}\!=\!\max_i \mathrm{VEG}_i$; $\mathrm{HitMax}\!\in\!\{0,1\}$ (whether the tool-call limit is hit); and $\mathrm{POER}\!\in\!\{0,1\}$ (Eq.~\ref{eq:poer}). Each feature targets one sub-case of \S\ref{sec:two_failure_modes}. With $\tau_{\text{sat}}\!=\!0.95$ and $\epsilon\!=\!0.01$, Algorithm~\ref{alg:diagnostic} assigns each trajectory $\tau$ to exactly one group.

\paragraph{Choice of features and thresholds.}
The diagnostic is not a learned classifier and does not use outcome labels to fit a decision boundary. It is a deterministic partition whose features are tied to the two failure definitions above. $g_0$ detects a pre-call ceiling effect; $V^{\max}$ checks whether any observation carries visual evidence; and $\mathrm{POER}$ with $\mathrm{HitMax}$ marks failures to stop after useful evidence has been exhausted. We set $\tau_{\text{sat}}\!=\!0.95$ because a probability gap above this value leaves at most a small improvement margin for a later call. We set $\epsilon\!=\!0.01$ because changes below one probability point are treated as near-zero visual contribution. Appendix~\ref{app:diagnostic-sensitivity} reports sensitivity checks over $\epsilon$ and $\tau_{\text{sat}}$ variants; the model-level ordering and the conclusion that positive ATE is concentrated in the Calibrated subset are unchanged.

\begin{algorithm}[t]
\caption{Diagnostic classification of a tool-use trajectory $\tau$.}
\label{alg:diagnostic}
\small
\begin{algorithmic}[1]
\Require Features $(n, g_0, V^{\max}, \mathrm{HitMax}, \mathrm{POER})$;
thresholds $\tau_{\text{sat}}\!=\!0.95$, $\epsilon\!=\!0.01$.
\Ensure Group label in
$\{\textsc{No-call},$ $\textsc{Mode 1 (CWL)},$
$\textsc{Mode 2 (LWP)},$ $\textsc{Calibrated}\}$.
\If{$n = 0$}
    \State \Return \textsc{No-call}
\EndIf
\If{$g_0 > \tau_{\text{sat}}$ \textbf{or} $V^{\max} < \epsilon$}
    \State \Return \textsc{Mode 1 (CWL)}
    \Comment{saturated prior or inactive call}
\EndIf
\If{$\mathrm{POER} = 1$ \textbf{or} $\mathrm{HitMax} = 1$}
    \State \Return \textsc{Mode 2 (LWP)}
    \Comment{informative call, planning failure}
\EndIf
\State \Return \textsc{Calibrated}
\end{algorithmic}
\end{algorithm}


\begin{table}[t]
\caption{\textbf{Per-model group distribution on V$^{\!*}$.} Rows sum to
$100\%$. Same $(\tau_{\text{sat}}, \epsilon)\!=\!(0.95, 0.01)$ for every
model.}
\label{tab:group_fractions}
\centering
\footnotesize
\setlength{\tabcolsep}{4pt}
\begin{tabular}{lcccc}
\toprule
Model & No-call & Mode 1 & Calibrated & Mode 2 \\
      &         & (CWL)  &            & (LWP)  \\
\midrule
DeepEyes        & $19.4$  & $72.8$ & $7.9$  & $0.0$ \\
Qwen3-VL-4B     & $13.1$ & $58.6$ & $24.1$ & $4.2$ \\
Qwen3-VL-8B     & $5.2$  & $70.7$ & $20.9$ & $3.1$ \\
Mini-o3         & $10.5$ & $34.6$ & $45.0$ & $9.9$  \\
\bottomrule
\end{tabular}
\end{table}
\paragraph{Per-model results.}
Table~\ref{tab:group_fractions} reports the group distribution on V$^{\!*}$, making the model-level prototypes precise. \textbf{DeepEyes} is a pure-Mode-1 prototype with essentially no Mode~2 mass. \textbf{Mini-o3} is the only Calibrated-leading prototype, paired with a substantial Mode~2 fraction (uniform per-call decay). The \textbf{Qwen3-VL} family exhibits three-group co-existence: Mode~1 dominates (saturated calls), with a non-trivial Calibrated group and a small Mode~2 tail (early saturation with continued calls).

\paragraph{Behavioral validation.}
We verify each label against independent rollout behavior. Mode~1 pre-call $\texttt{<think>}$ blocks already commit to the final $Y$ in $\sim\!58\%$ of cases (vs.\ $\sim\!28\%$ for Calibrated): the calls are performative rather than perceptual, echoing unfaithfulness reported by \citet{xu2025defacto,hou2025codev}. Mode~2 trajectories on Mini-o3 often zoom into tiny regions ($38\%$ of crops cover $<\!4\%$ of $I$) and abandon a crop for an unrelated one rather than refining; on Qwen3-VL-8B the model commits internally early yet keeps issuing zoom calls. Details in App.~\ref{app:behavioral-validity}.

\subsection{Decomposition of the Policy-Level ATE}
\label{sec:ate_decomposition}

We use the resulting group distribution to decompose the policy-level ATE into per-group contributions. Let $f_b$ be the fraction of trajectories assigned to group $b$ and $\bar\Delta_{\text{policy},b}$ be the mean accuracy gap between $\pi_{\text{tool}}$ and $\pi_{\text{direct}}$ on those trajectories:
\begin{equation}
\label{eq:ate_decomp_main}
\text{ATE}_{\text{policy}} \;=\; \!\!\sum_{b\in\mathcal{B}} f_b \,\bar\Delta_{\text{policy},b},
\end{equation}
with $\mathcal{B}\!=\!\{\textsc{NoCall},\textsc{M1},\textsc{Cal},\textsc{M2}\}$. Table~\ref{tab:ate_decomp} realises this on V$^{\!*}$ for the four interface-based models with full diagnostic coverage (full breakdown in App.~\ref{app:ate-decomposition}).

\begin{table}[t]
\caption{\textbf{Group-wise decomposition of the policy-level ATE on
V$^{\!*}$.} Each cell is the contribution
$f_b\!\cdot\!\bar\Delta_{\text{policy},b}$ in pp.
The last column is computed from unrounded contributions and matches
Table~\ref{tab:policy_ate}; displayed components may not sum exactly due to rounding.
$(\tau_{\text{sat}},\epsilon)\!=\!(0.95, 0.01)$.}
\label{tab:ate_decomp}
\centering
\footnotesize
\setlength{\tabcolsep}{3pt}
\begin{tabular}{lccccc}
\toprule
Model & No-call & Mode 1 & Cal. & Mode 2 & $\text{ATE}_{\text{policy}}$ \\
\midrule
DeepEyes     & $0.0$  & $-0.5$ & $+0.5$ & $0.0$  & $0.0$  \\
Qwen3-VL-4B  & $-0.2$ & $+2.0$ & $+3.5$ & $-0.9$ & $+4.4$ \\
Qwen3-VL-8B  & $-0.5$ & $-0.2$ & $+7.6$ & $-0.1$ & $+6.9$ \\
Mini-o3      & $-0.3$ & $+1.9$ & $+3.3$ & $+0.6$ & $+5.5$ \\
\bottomrule
\end{tabular}
\end{table}

\paragraph{Calibrated carries the ATE.}
\emph{Calibrated} is the only column positive across all four models, and accounts for the majority of every non-zero ATE under the default diagnostic: only on this subset does $T\!\to\!O\!\to\!Y$ deliver a positive effect at population scale. Mode~1 is small or offsetting, and its non-zero entries arise because $g_0$ is read on the tool-use rollout while the direct answer is produced by an independent no-tool rollout. Mode~2 and No-call stay within $\sim\!\pm 1$\,pp. The same qualitative pattern is reproduced on HR-Bench-4K, where the Calibrated subset again carries most of the positive ATE (App.~\ref{app:hrbench-validation}).

\section{Discussion}
\label{sec:discussion}
\paragraph{The illusion of visual tool-use.}
\label{sec:the-illusion-of-visual-tool-use}
The aggregate policy-level accuracy gain from visual tool-use is real, but it can create the misleading impression that tool-use is broadly and causally effective across rollouts. Our causal audit shows instead that the gain is concentrated in a calibrated minority, while many other tool-using trajectories either fail to use the returned visual evidence or invoke tools with an incoherent schedule. We call this phenomenon \emph{the illusion of visual tool-use}: \textbf{\textit{Having} a tool does not necessarily mean it is \textit{used} properly}. The question ``does the model think with images?'' is therefore more important than ``does the model call visual tools?''. Current policies satisfy the latter far more often than the former. Improving tool-use capability should thus focus not only on richer interfaces and longer rollouts, but also on the causal coupling between actions, the evidence they elicit, and the final answer.

\paragraph{An RL-trap hypothesis.}
\label{sec:rl-trap}
Mode~1 and Mode~2 appear across models differing in architecture, data, and tool interface; a plausible common factor is outcome-only RL over tool-augmented rollouts. We hypothesize that this objective populates \emph{both} failure groups: rewarding tool-calling behavior reinforces the $T\!\to\!Y$ shortcut (Mode~1); the absence of any penalty on redundant or harmful intermediate steps along correct trajectories rewards Mode~2; and a Calibrated and a Mode~1 trajectory that both reach the correct answer receive identical reward, so the objective cannot prefer the former. Concurrent process-aware training work is consistent with this view: outcome-only rewards cannot distinguish useful, redundant, and misleading tool calls~\citep{feng2026taco,wang2025virl}. Establishing the mechanism would require matched training that varies only the reward signal, which we leave as future work.

\paragraph{Operational uses.}
\label{sec:diagnostic-operational}
The diagnostic is actionable. \textbf{(i)~Beyond-accuracy evaluation}: any intervention is graded by the distributional shift $\Delta f_{\text{Cal}}\!>\!0$ and $\Delta f_{\text{M1}}\!+\!\Delta f_{\text{M2}}\!<\!0$, regardless of whether $\text{ATE}_{\text{policy}}$ moves. \textbf{(ii)~Inference-time adaptation}: Offline diagnostic can motivate/train an inference-time gating or stopping policy. Mode~1 trajectories can bypass tool-use; Mode~2 trajectories call for an early-stopping rule that commits before late-stage harmful calls. \textbf{(iii)~Process-aware credit assignment}: step-level VEG supplies per-step supervision missing from outcome-only reward, up-weighting Calibrated behavior while penalizing wasted Mode~1 and harmful Mode~2 calls.

\section{Conclusion}

We revisited the thinking-with-images paradigm through a causal lens, separating \emph{observation-mediated paths} from \emph{action-induced shortcuts} and operationalizing this distinction through interventions at the policy, trajectory, and step levels. Across six models and five benchmarks, our audit identifies \emph{policy miscalibration} as a central bottleneck: models either call tools without causally using the returned evidence (\emph{Calling Without Looking}) or use that evidence under a poorly planned calling strategy (\emph{Looking Without Planning}). Our additive decomposition further shows that aggregate accuracy gains are carried largely by a \emph{Calibrated} minority. We refer to this discrepancy between aggregate gains and trajectory-level effectiveness as \emph{the illusion of visual tool-use}: benchmark-level improvements make visual tool-use appear broadly effective, even though many tool-using trajectories are not causally grounded in the observations they request or do not invoke them coherently. These findings suggest that progress should be measured not only by final accuracy or tool-call frequency, but by whether models call tools when needed, use the returned evidence, and translate it into better answers.

\section*{Limitations}

Our study has several limitations that constrain the scope of its conclusions and point to directions for future work.

\paragraph{Model access constraints.} All experiments are conducted on open-source thinking-with-images
models. Our intervention levels have different access requirements: the policy-level and trajectory-level interventions only require running the model and controlling the returned observations, so they can apply to black-box models with an observable tool interface, whereas the step-level VEG uses token-level scores and thus needs white-box access. Closed-source models such as OpenAI o3/o4-mini \citep{openai2025o3o4mini} expose neither their token-level probabilities nor the ability to intervene on intermediate observations, so whether the same conclusions extend to the closed-source models is an open question that we cannot resolve here.

\paragraph{Limited tool set.} We only focus on the \textsc{crop-and-zoom} operation, which is the dominant tool in current thinking-with-images models.  Other tools such as image segmentation, OCR, frame selection in video, code-based image manipulation, or external search may exhibit qualitatively different calibration behaviors, but our causal framework transfers by constructing a tool-specific counterfactual observation. For OCR one can replace the returned text with text from another region or an empty result; for localization one can replace the returned box with a shifted or random box; for video frame selection one can replace the chosen frame with a temporally shifted or irrelevant frame. We do not claim the crop-and-zoom patterns necessarily hold for these tools.

\paragraph{RL trap as a hypothesis.} The \emph{RL-trap} account in \S\ref{sec:rl-trap}, which attributes
policy miscalibration to outcome-only reinforcement learning over tool-augmented rollouts, is a hypothesis consistent with our observations rather than a causal claim established by training-time experiments. Our empirical findings, the two failure modes and the concentration of gains in the Calibrated subset, hold regardless of whether this hypothesis is confirmed. Concurrent process-aware training work provides external evidence consistent with it~\citep{feng2026taco,wang2025virl}. Definitively isolating the role of outcome-only rewards
would require controlled training studies, e.g.\ matched runs that
vary only the reward signal (outcome-only vs.\ process-aware rewards), the rollout policy (tool-augmented vs.\
direct), or the credit-assignment scheme, while holding data and
backbone fixed. We view such controlled training as an important
follow-up but beyond the diagnostic scope of this paper.

\section*{Acknowledgments}

This work is supported by the Shanghai Artificial Intelligence Laboratory.

\bibliography{custom}

\appendix

\section{Experimental Setup Details}
\label{app:experimental_setup}
\label{sec:appendix}

This appendix specifies the experimental pipeline behind
\S\ref{sec:policy_level_intervention}--\S\ref{sec:step_level_intervention}:
the six thinking-with-images policies and their decoding configurations
(\S\ref{app:setup-models}), the seven evaluation benchmarks and the dual-layer
answer verification protocol (\S\ref{app:setup-benchmarks}), the implementation
of each of the three causal interventions
(\S\ref{app:setup-interventions}), and the aggregated compute budget
(\S\ref{app:setup-compute}).

\subsection{Models and Inference Configuration}
\label{app:setup-models}

\paragraph{Model checkpoints.}
We evaluate the six publicly released thinking-with-images models listed in
Table~\ref{tab:app-checkpoints}. Five of them invoke image operations through a predefined \textsc{crop-and-zoom} tool with a fixed action schema (\emph{interface-based}); Thyme instead exposes a Python sandbox, where the same crop-and-zoom operations are carried out by model-generated Python code (\emph{code-based}). For every model
we use the officially released checkpoint without additional fine-tuning,
adapters, or quantization.

\begin{table*}[t]
\caption{\textbf{Model checkpoints.} HF Repo lists the Hugging Face identifier of the checkpoint used.}
\label{tab:app-checkpoints}
\centering
\small
\setlength{\tabcolsep}{4pt}
\begin{tabular}{l l l l}
\toprule
Model & Backbone & Tool interface & HF Repo \\
\midrule
DeepEyes~\citep{zheng2025deepeyes}        & Qwen2.5-VL-7B  & Interface & \texttt{ChenShawn/DeepEyes-7B} \\
Pixel Reasoner~\citep{wang2025pixel}      & Qwen2.5-VL-7B  & Interface     & \texttt{TIGER-Lab/PixelReasoner-RL-v1} \\
Mini-o3~\citep{lai2025mini-o3}            & Qwen2.5-VL-7B  & Interface & \texttt{Mini-o3/Mini-o3-7B-v1} \\
Qwen3-VL-8B~\citep{bai2025qwen3vl}        & Qwen3-VL-8B    & Interface & \texttt{Qwen/Qwen3-VL-8B-Instruct} \\
Qwen3-VL-4B~\citep{bai2025qwen3vl}        & Qwen3-VL-4B    & Interface & \texttt{Qwen/Qwen3-VL-4B-Instruct} \\
Thyme~\citep{zhang2026thyme}              & Qwen2.5-VL-7B  & Code              & \texttt{Kwai-Keye/Thyme-RL} \\
\bottomrule
\end{tabular}
\end{table*}

\paragraph{System prompts.}
The policy-level intervention is realized at the system-prompt level, with
model weights and decoding parameters held fixed across the two inference modes. In the
tool-use rollout, each model uses the system prompt and tool-call pipeline provided
with its official release, including the action schema, parser, special tokens,
and return format. In the direct rollout, we replace the model's system prompt with
the following tool-free instruction:
\begin{quote}
\texttt{You are a helpful assistant. Please reason step by step and put your
answer within \textbackslash boxed\{\}.}
\end{quote}
We also disable the tool-call pipeline so that any incidentally emitted tool
tokens are treated as plain text. Both rollouts decode under the same
generation configuration in Table~\ref{tab:app-decoding}.

\begin{table*}[t]
\caption{\textbf{Per-model decoding configuration.} Values follow each model's
official inference configuration. Both the tool-use and direct rollouts use the same
configuration. ``Greedy'' indicates a deterministic setting. $T_{\max}$ is the
tool-call limit.}
\label{tab:app-decoding}
\centering
\footnotesize
\setlength{\tabcolsep}{4pt}
\begin{tabular}{l cccc c}
\toprule
Model & Temp. & top-$p$ & top-$k$ & Mode & $T_{\max}$ \\
\midrule
DeepEyes        & $0.0$  & $1.0$   & $-1$ & Greedy     & $12$ \\
Pixel Reasoner  & $0.1$  & $0.95$  & $-1$ & Stochastic & $12$ \\
Mini-o3         & $1.0$  & $1.0$   & $-1$ & Stochastic & $12$ \\
Qwen3-VL-8B     & $0.7$  & $0.8$   & $20$ & Stochastic & $12$ \\
Qwen3-VL-4B     & $0.7$  & $0.8$   & $20$ & Stochastic & $12$ \\
Thyme           & $0.01$ & $0.001$ & $1$  & Greedy     & $12$ \\
\bottomrule
\end{tabular}
\end{table*}

\paragraph{Serving stack.}
All models are served with \texttt{vLLM} using bf16 weights. Image inputs are
processed by each model's native preprocessor. A rollout server is allocated one
NVIDIA H200 GPU.

\subsection{Benchmarks and Answer Verification}
\label{app:setup-benchmarks}

We evaluate seven benchmark partitions from four benchmark families. The main
text reports VisualProbe as a single aggregate column over Easy, Medium, and
Hard.

\begin{table*}[t]
\caption{\textbf{Benchmarks evaluated.} MCQ denotes multiple choice and OE
denotes open-ended free-form answering.}
\label{tab:app-benchmarks}
\centering
\footnotesize
\setlength{\tabcolsep}{5pt}
\begin{tabular}{l l l c}
\toprule
Benchmark & Source & Format & Size \\
\midrule
V$^{\!*}$ & \citet{wu2024v} & MCQ & $191$ \\
HR-Bench-4K & \citet{wang2025hrbench} & MCQ & $800$ \\
HR-Bench-8K & \citet{wang2025hrbench} & MCQ & $800$ \\
VisualProbe-Easy & \citet{lai2025mini-o3} & OE & $141$ \\
VisualProbe-Medium & \citet{lai2025mini-o3} & OE & $268$ \\
VisualProbe-Hard & \citet{lai2025mini-o3} & OE & $106$ \\
MME-RealWorld-Lite & \citet{zhang2025mmerealworld} & MCQ & $1919$ \\
\bottomrule
\end{tabular}
\end{table*}

In free-form generation, a model's output is rarely an exact match to the gold answer. So we
verify correctness with a two-stage pipeline. First, we extract
the final answer from the last \texttt{<answer>...</answer>} tag in the
response, or from the last \texttt{\textbackslash boxed\{...\}} expression when
no answer tag is present. We strip whitespace and trailing punctuation,
lowercase the string, and collapse MCQ responses to the option letter when one
is present. The normalized answer is compared against the normalized gold answer.

When exact match fails, we deploy Qwen3-30B-A3B-Instruct-2507~\citep{bai2025qwen3vl}
as a judge. The judge receives the question, the gold answer, and
the model response, then outputs a \texttt{correct}/\texttt{incorrect} label. The
judge is decoded greedily with thinking mode disabled. A response is scored as
correct if either the exact-match stage succeeds or the judge returns
\texttt{correct}.

\subsection{Intervention Implementations}
\label{app:setup-interventions}

\paragraph{Policy-level intervention (\S\ref{sec:policy_level_intervention}).}
Implemented by substituting the system prompt as detailed in \S\ref{app:setup-models}; all other components (decoding parameters, tool engine, evaluation protocol) remain identical across the two modes.

\paragraph{Trajectory-level intervention (\S\ref{sec:trajectory_level_intervention}).}
At every step $i$ we let the policy emit $T_i$ as usual but replace the engine
output $O_i = E_{\text{tool}}(T_i, I)$  with a corrupted observation $\tilde O_i$ with the same shape as $O_i$. The counterfactual is generated by one of three corruption schemes:
\begin{itemize}[leftmargin=*,itemsep=2pt,topsep=2pt]
\item \textsc{random-crop} (standard, used in main-text
Table~\ref{tab:traj_results}): a same-shape crop sampled uniformly at random
from the source image $I$, preserving natural-image statistics.
\item \textsc{noise} (Appendix~\ref{app:traj_corruption}): a same-sized RGB image whose pixels are sampled i.i.d. from a per-channel discrete uniform distribution over $[0, 255]$.
\item \textsc{blank} (Appendix~\ref{app:traj_corruption}): a same-shape all-zero (pure black) RGB image.
\end{itemize}
Trajectory rollouts under each corruption scheme respect the same tool call limit $T_{\max}$ as the vanilla rollouts (Table~\ref{tab:app-decoding}); a trajectory
that emits tool calls beyond the limit without committing a final answer is scored as incorrect.

We considered replaying original trajectories while replacing the returned images, but this setup cannot cleanly remove visual evidence. In thinking-with-images models, natural-language thoughts and tool actions are interleaved, and each thought is conditioned on preceding observations. Replaying the original thoughts would therefore leak information from clean observations into later decisions. Replaying only the action coordinates avoids this leakage, but prevents the model from adapting its subsequent actions to the corrupted observations. We therefore use online corruption, which presents corrupted observations during the rollout and prevents later thoughts and actions from inheriting information from clean returned images. Replay-style analysis remains useful as a complementary stress test, but not as a leakage-free causal diagnostic.

\paragraph{Step-level intervention (\S\ref{sec:step_level_intervention}).}
The counterfactual observation uses the same \textsc{random-crop} construction as the trajectory-level intervention. Each VEG value is obtained by averaging over three independently sampled counterfactual crops.
For the answer readout, we do not run an extra free-form answer rollout.
Instead, at each checkpoint we keep the generated prefix through the current thinking block, append the literal string \texttt{<answer>}, and read the next-token logits.
We then apply an option-restricted softmax over the candidate labels and compute the probability gap in Eq.~\ref{eq:prob_gap}.
The same readout is applied before the call, after the real observation, and after the counterfactual observation, with the prefix and tool action held fixed across the real/counterfactual pair.
VEG is therefore a local counterfactual estimate of observation-mediated influence under this fixed readout, not a claim that we recover the full effect of the tool call on unconstrained final-answer formation.

However, the readout is not detached from natural decoding: the logits are exactly the distribution from which the model selects its next token when it naturally emits an answer label after \texttt{<answer>}. A free-form rollout ultimately exposes only a discrete outcome, namely whether the final answer is correct or incorrect. In contrast, the option-restricted logits preserve the graded strength of the model's answer preference, allowing us to quantify whether it is strongly certain, weakly uncertain, or near-indifferent between the correct option and alternatives at the same reasoning state. Its purpose is therefore to test whether the returned visual content changes this immediate, decoding-relevant answer preference. We note that similar logit-based belief-update signals have also been used in recent agent-training work \citep{wang2026igpo}.

\subsection{Compute Resources}
\label{app:setup-compute}

The experimental pipeline runs on NVIDIA H200 GPUs. The inference of each model needs at least one GPU, and we use eight GPUs in total during all experiments. The judge model is served
separately with \texttt{vLLM}, bf16 weights, and tensor parallelism across two H200 GPUs.

\section{Additional Trajectory-Level Observation Corruptions}
\label{app:traj_corruption}

This appendix complements \S\ref{sec:trajectory_level_intervention} along five axes that the main text could not accommodate due to space:
\textbf{(i)}~the two corruption schemes deferred from the main \textsc{random-crop}
table (\textsc{noise} and \textsc{blank}) reported on V$^{\!*}$ across all six models  (\S\ref{app:corrupt-vstar});
\textbf{(ii)}~trajectory-length statistics under intervention, contrasting the
mean number of tool calls per item between the vanilla and intervened
conditions (\S\ref{app:corrupt-callcount}); \textbf{(iii)}~several qualitative regenerated $\texttt{<think>}$
blocks under each corruption scheme (\S\ref{app:corrupt-qualitative}); \textbf{(iv)}~a force-answer ablation experiment to separate the effect of evidence loss from truncation (\S\ref{app:force-answer}); and \textbf{(v)}~results of repeated-seed runs to ensure the robustness of the results.
All experiments use the same protocol, models, and benchmarks as
Table~\ref{tab:traj_results}.

\subsection{V* Corruption-Choice Ablation}
\label{app:corrupt-vstar}

Table~\ref{tab:app-traj-vstar-corruption} compares
\textsc{random-crop}, \textsc{noise}, and \textsc{blank} on V$^{\!*}$, which
checks whether the trajectory-level conclusions are sensitive to the corruption
choice. The key qualitative finding is that the Qwen3-VL family shows a
markedly smaller drop under \textsc{blank} than under \textsc{random-crop}:
the policy detects corruption but is fooled by statistically plausible ones.

\begin{table*}[t]

\caption{\textbf{Trajectory-level corruption-choice ablation on V$^{\!*}$.}
Each cell reports accuracy under the dynamic intervention, the corresponding
$\Delta_{\text{traj}}$ in pp, and Hit-MaxT.}
\label{tab:app-traj-vstar-corruption}
\centering
\small
\setlength{\tabcolsep}{3pt}
\begin{tabular}{l ccc ccc ccc}
\toprule
Model
& \multicolumn{3}{c}{\textsc{random-crop}}
& \multicolumn{3}{c}{\textsc{noise}}
& \multicolumn{3}{c}{\textsc{blank}} \\
\cmidrule(lr){2-4}\cmidrule(lr){5-7}\cmidrule(lr){8-10}
& Acc & $\Delta$ & Hit
& Acc & $\Delta$ & Hit
& Acc & $\Delta$ & Hit \\
\midrule
DeepEyes        & 83.8 & +0.5 & 0\% & 83.8 & +0.5 & 0\% & 82.9 & -0.4 & 0\% \\
Pixel Reasoner  & 82.7 & -2.1 & 0\% & 82.2 & -2.6 & 0\% & 83.8 & -1.0 & 0\% \\
Mini-o3         & 23.6 & -64.2 & 84.8\% & 20.4 & -67.4 & 78.1\% & 13.6 & -74.2 & 84.8\% \\
Qwen3-VL-8B     & 30.4 & -60.7 & 59.2\% & 28.8 & -62.3 & 64.4\% & 61.8 & -29.3 & 26.2\% \\
Qwen3-VL-4B     & 38.2 & -48.2 & 50.3\% & 48.7 & -37.7 & 38.7\% & 53.9 & -32.5 & 37.7\% \\
Thyme           & 82.2 & -1.0 & 0\% & 82.2 & -1.0 & 0\% & 81.7 & -1.5 & 0\% \\
\bottomrule
\end{tabular}
\end{table*}

\begin{table*}[t]
\centering
\captionof{table}{\textbf{Mean number of tool calls per item on V$^{\!*}$
under intervention.} Contrasts the vanilla baseline with each corruption
scheme.}
\label{tab:app-traj-callcount-vstar}
\small
\setlength{\tabcolsep}{3pt}
\begin{tabular}{l cccc}
\toprule
Model & Vanilla & \textsc{r-crop} & \textsc{noise} & \textsc{blank} \\
\midrule
DeepEyes       & 1.0 & 1.0 & 1.0 & 1.0 \\
Pixel Reasoner & 0.9   & 0.9 & 0.9  & 0.9 \\
Mini-o3        & 2.1 & 9.1 & 10.1 & 9.6 \\
Qwen3-VL-8B    & 1.1 & 6.5 & 7.2 & 4.5 \\
Qwen3-VL-4B    & 1.1   & 6.5 & 5.1  & 4.9 \\
Thyme          & 0.1   & 0.1 & 0.1  & 0.1 \\
\bottomrule
\end{tabular}
\end{table*}

\subsection{Trajectory Length under Intervention}
\label{app:corrupt-callcount}

A complementary diagnostic to Hit-MaxT is the \emph{mean number of tool calls
per item} in the intervened condition, contrasted with the vanilla baseline.
We report the results on V$^{\!*}$ across six models in Table~\ref{tab:app-traj-callcount-vstar}. Notably, the growth in average number of tool calls under observation intervention is highly consistent with the fraction of trajectories reaching the tool-call limit.

\subsection{Qualitative Examples of Regenerated Think Blocks}
\label{app:corrupt-qualitative}

To complement the quantitative analysis, we provide three qualitative samples
of regenerated $\texttt{<think>}$ blocks under each corruption
scheme (the question is ``What is the material of the glove? A. rubber B. cotton C. kevlar D. leather''):
\begin{itemize}[leftmargin=*,itemsep=2pt,topsep=2pt]
\item \textbf{Mini-o3 under \textsc{random-crop}:} the model realizes that the received crop is not consistent with the expectation and retries with a new bounding box on the full image.
\begin{grayquote}\noindent
\texttt{<think>}
The user wants to know the material of the glove in the image.
The previous action was to select a bounding box that I thought contained the glove, but it seems I missed it.
Let's re-examine the original image.
I can see a person in the upper part of the image.
I will try a different bounding box to get a closer view of the person's hand and glove.
I'll select the region containing the person wearing the glove from the original image for a better view.
\texttt{</think>}

\texttt{<tool\_call>} ... \texttt{</tool\_call>}
\end{grayquote}
\item \textbf{Qwen3-VL-8B under \textsc{blank}:} the model identifies the
blank patch as a tool failure and continues its reasoning based on the original image.
\begin{grayquote}\noindent
I apologize for the technical issue with the image display. Based on the visual information available, I can still analyze the glove's appearance.

Looking at the glove in the original image, it is bright blue, thin, and appears to be made of a flexible material that conforms to the hand. This is typical of disposable food service gloves. In food preparation environments, such gloves are most commonly made of **nitrile** or **vinyl**, which are types of synthetic rubber. While ``rubber'' is not technically the same as nitrile or vinyl, it is the closest option among the given choices, as these materials are often referred to as ``rubber gloves'' in common language.

Now, reviewing the findings: The glove is thin, flexible, and used in a food handling context, which strongly suggests it is made of a synthetic rubber material.

Therefore, the correct answer is:

\texttt{<answer>} A \texttt{</answer>}
\end{grayquote}
\item \textbf{DeepEyes under \textsc{noise}:} the model commits an answer
without acknowledging the corrupted observation.
\begin{grayquote}\noindent
Turn 1: \texttt{<think>}
The glove in the image appears to be made of a flexible material that is commonly used for food handling. Given the options provided, rubber is a common material for gloves used in food preparation due to its flexibility and ability to provide a good grip. Let's zoom in on the glove to confirm its material. \texttt{</think>}

\texttt{<tool\_call>} ... \texttt{</tool\_call>}

Turn 2: \texttt{<think>} The glove in the image is indeed made of a flexible material that matches the appearance of rubber. The glove is blue and has a smooth, pliable texture, which is characteristic of rubber gloves used in food handling. Therefore, the correct answer is A. rubber. \texttt{</think>}

\texttt{<answer>} A. rubber \texttt{</answer>}

\end{grayquote}
\end{itemize}

\subsection{Force-Answer Ablation Experiment: Separating Evidence Loss from Truncation}
\label{app:force-answer}

The trajectory-level corruption in \S\ref{sec:trajectory_level_intervention} mixes two effects: the loss of useful visual observations, and induced repair loops that may be truncated at the tool-call budget without committing an answer. To separate them, we rerun the \textsc{random-crop} corruption on V$^{\!*}$ with a \emph{force-answer} rule: when a rollout reaches the tool-call limit, we tell the model that no more tools are allowed and force it to produce a final answer immediately. This preserves the corrupted-observation history while removing failures caused only by having no answer at truncation.

\begin{table}[t]
\caption{\textbf{Force-answer Experiment on V$^{\!*}$ under \textsc{random-crop}.}
\emph{Clean} is tool-use accuracy; \emph{Orig.\ corr.} is the original corrupted accuracy (Table~\ref{tab:traj_results}); \emph{Force-ans.} forces an answer at the tool-call limit; \emph{Rem.\ drop} is the remaining drop of the force-answer run relative to clean (pp).}
\label{tab:app-force-answer}
\centering
\small
\setlength{\tabcolsep}{4pt}
\begin{tabular}{lcccc}
\toprule
Model & Clean & Orig.\ corr. & Force-ans. & Rem.\ drop \\
\midrule
Mini-o3     & $87.8$ & $23.6$ & $36.6$ & $51.2$ \\
Qwen3-VL-8B & $91.1$ & $30.4$ & $73.0$ & $18.1$ \\
Qwen3-VL-4B & $86.4$ & $38.2$ & $69.6$ & $16.8$ \\
\bottomrule
\end{tabular}
\end{table}

Table~\ref{tab:app-force-answer} shows that forced answering recovers a large part of the original drop for Qwen3-VL (from $30.4/38.2$ to $73.0/69.6$), so truncation of repair loops explains part of the degradation. However, all force-answer accuracies remain well below clean tool-use accuracy, by $18.1$ pp for Qwen3-VL-8B, $16.8$ pp for Qwen3-VL-4B, and $51.2$ pp for Mini-o3. Useful visual evidence is therefore genuinely lost under corrupted observations, and Mini-o3 remains especially brittle even with the forced-answer instruction. This supports reading the main trajectory-level result as the total effect of corrupting the observation-feedback process, which includes both evidence removal and induced repair behavior.

\subsection{Results of Repeated Runs}
\label{app:multiseed}

A single stochastic rollout does not characterize the variability of the trajectory-level results. We therefore repeat the V$^{\!*}$ random-crop analysis with independent random seeds while holding the checkpoint, prompt, decoding configuration, benchmark items, and scoring protocol fixed. The seed controls both stochastic decoding and the sampling of corrupted crops.

The original and two additional random-crop runs give corrupted accuracies of $23.6\%$, $22.5\%$, and $24.6\%$ for Mini-o3; $30.4\%$, $36.1\%$, and $28.8\%$ for Qwen3-VL-8B; and $38.2\%$, $46.1\%$, and $42.9\%$ for Qwen3-VL-4B. The exact value varies, but all runs retain the large degradation relative to clean tool-use.

\section{Additional Analysis of Step-Level Intervention on MCQ Benchmarks}
\label{app:veg_full}

This appendix provides more results and analysis of the step-level intervention:
(\S\ref{sec:step_level_intervention})
\textbf{(i)}~repeated-run results that characterize uncertainty across stochastic rollouts
(\S\ref{app:step-multiseed}); \textbf{(ii)}~per-position VEG details on
V$^{\!*}$ that support the within-trajectory dynamics.
(\S\ref{app:per_pos_detail}); \textbf{(iii)}~a concept-level comparison
clarifying how POER differs from the raw saturation rate
(\S\ref{app:poer-vs-saturation}).
The open-ended extension is deferred to Appendix~\ref{app:veg_openended}.

\subsection{Results of Repeated Runs and Uncertainty}
\label{app:step-multiseed}

A single stochastic rollout does not characterize the variability of the step-level results. We therefore repeat the V$^{\!*}$ analysis with independent random seeds while holding the checkpoint, prompt, decoding configuration, benchmark items, and scoring protocol fixed. The seed controls both stochastic decoding and the sampling of counterfactual crops.

Over four seeds, Mini-o3 has tool-use accuracy $85.1\%\!\pm\!1.8$ pp and mean VEG $0.0629\!\pm\!0.0041$, and Qwen3-VL-8B has $90.6\%\!\pm\!0.6$ pp and $0.1944\!\pm\!0.0183$ (mean $\pm$ sample standard deviation). The correct/incorrect split is stable in every run: Mini-o3 is positive on correct trajectories ($0.078$--$0.092$) and negative on incorrect ones ($-0.032$--$-0.006$); Qwen3-VL-8B is positive on correct ($0.237$--$0.306$) and negative on incorrect ($-0.268$--$-0.083$). The evidence-use patterns that motivate our diagnostic are thus not driven by one decoding seed.

\subsection{Per-Position VEG Detail}
\label{app:per_pos_detail}

This appendix expands the within-trajectory analysis of \S\ref{sec:step_level_intervention} along two axes that we omitted from the main text for space. Table~\ref{tab:vmax_split} reports the per-trajectory peak $V^{\max}\!=\!\max_i \mathrm{VEG}_i$ split by trajectory correctness (the pooled column reproduces the rightmost column of Table~\ref{tab:veg_results}); we also report the per-trajectory \emph{trough} $V^{\min}\!=\!\min_i \mathrm{VEG}_i$, which complements the harmful-rate by quantifying worst-case per-trajectory damage. Table~\ref{tab:per_pos_detail} expands Fig.~\ref{fig:per_pos_dynamics} with the wasted-rate ($\Pr[|\mathrm{VEG}|\!<\!0.01]$) and harmful-rate ($\Pr[\mathrm{VEG}\!<\!-0.05]$) at each position bin, reported both pooled over all calls and restricted to the non-saturated subset ($g_{i-1}\!\le\!\tau_{\text{sat}}\!=\!0.95$); the per-position saturation rate is reported alongside. Throughout this appendix we use the \emph{first-of-turn} convention: when a turn emits multiple parallel tool calls, only the first call is counted and $k$ is indexed by turn. This matches the turn-level definition used by POER. Under this convention, DeepEyes never produces a second turn of tool calls on V$^{\!*}$ ($0/191$ trajectories; the only $2/191$ multi-call cases are parallel calls collapsed within the first turn), so all $k\!\geq\!2$ cells are empty by construction; correspondingly, $\mathrm{POER}$ (Eq.~\ref{eq:poer}) is undefined for DeepEyes because the conditioning event $\{k\!\geq\!2,\, g_0\!\le\!\tau_{\text{sat}}\}$ has no support, as flagged in Fig.~\ref{fig:per_pos_dynamics}(b) and discussed in \S\ref{sec:step_level_intervention}. Bins with very small $n$ are flagged.

\begin{table*}[t]
\caption{\textbf{Per-trajectory peak/trough VEG on V$^{\!*}$} (mean over trajectories). Pooled (\textbf{A}) and split by trajectory correctness (\textbf{C}/\textbf{I}).}
\label{tab:vmax_split}
\centering
\small
\setlength{\tabcolsep}{6pt}
\begin{tabular}{l ccc ccc}
\toprule
& \multicolumn{3}{c}{mean $V^{\max}$} & \multicolumn{3}{c}{mean $V^{\min}$} \\
\cmidrule(lr){2-4}\cmidrule(lr){5-7}
Model & A & C & I & A & C & I \\
\midrule
DeepEyes    & $+0.05$ & $+0.07$ & $-0.01$ & $+0.05$ & $+0.07$ & $-0.01$ \\
Mini-o3     & $+0.19$ & $+0.21$ & $+0.10$ & $+0.04$ & $+0.08$ & $-0.16$ \\
Qwen3-VL-8B & $+0.30$ & $+0.37$ & $-0.43$ & $+0.14$ & $+0.21$ & $-0.63$ \\
\bottomrule
\end{tabular}
\end{table*}

\begin{table*}[t]

\caption{\textbf{Per-position VEG breakdown on V$^{\!*}$.} For each model and bin $k\!\in\!\{1,2,3,\geq\!4\}$ we report the number of calls $n$, mean VEG, fraction of calls with probability gap already past $\tau_{\text{sat}}$ (\emph{Sat}), the wasted-rate ($w$), and the harmful-rate ($h$). Each metric is given (a)~over all calls and (b)~restricted to non-saturated calls. ``--'' marks bins with $n\!=\!0$.}
\label{tab:per_pos_detail}
\centering
\small
\setlength{\tabcolsep}{3pt}
\begin{tabular}{l c r r r r r r r r}
\toprule
& & & \multicolumn{3}{c}{All calls} & & \multicolumn{3}{c}{Non-sat ($g_{i-1}\!\le\!\tau_{\text{sat}}$)} \\
\cmidrule(lr){4-6}\cmidrule(lr){8-10}
Model & $k$ & Sat & $n$ & mean VEG & $w$ \;/\; $h$ & & $n$ & mean VEG & $w$ \;/\; $h$ \\
\midrule
\multirow{4}{*}{DeepEyes}
 & $1$                & $36.4\%$ & $154$ & $+0.054$ & $81.2\%$ / $5.2\%$ & & $98$ & $+0.096$ & $73.5\%$ / $7.1\%$ \\
 & $2$  & --       & $0$   & --       & -- / --            & & $0$  & --       & -- / --            \\
 & $3$                & --       & $0$   & --       & -- / --            & & $0$  & --       & -- / --            \\
 & $\geq\!4$          & --       & $0$   & --       & -- / --            & & $0$  & --       & -- / --            \\
\midrule
\multirow{4}{*}{Mini-o3}
 & $1$           & $22.8\%$ & $171$ & $+0.104$ & $30.4\%$ / $11.1\%$ & & $132$ & $+0.135$ & $16.7\%$ / $14.4\%$ \\
 & $2$           & $25.0\%$ & $100$ & $+0.070$ & $29.0\%$ / $16.0\%$ & & $75$  & $+0.098$ & $16.0\%$ / $17.3\%$ \\
 & $3$           & $23.8\%$ & $42$  & $+0.032$ & $31.0\%$ / $19.0\%$ & & $32$  & $+0.041$ & $18.8\%$ / $21.9\%$ \\
 & $\geq\!4$     & $12.7\%$ & $126$ & $+0.011$ & $27.0\%$ / $20.6\%$ & & $110$ & $+0.015$ & $21.8\%$ / $21.8\%$ \\
\midrule
\multirow{4}{*}{Qwen3-VL-8B}
 & $1$           & $69.1\%$ & $181$ & $+0.196$ & $59.1\%$ / $8.3\%$  & & $56$ & $+0.632$ & $10.7\%$ / $19.6\%$ \\
 & $2$           & $56.6\%$ & $53$  & $+0.273$ & $47.2\%$ / $5.7\%$  & & $23$ & $+0.569$ & $8.7\%$  / $8.7\%$  \\
 & $3$           & $46.2\%$ & $13$  & $+0.271$ & $46.2\%$ / $0.0\%$  & & $7$  & $+0.500$ & $28.6\%$ / $0.0\%$  \\
 & $\geq\!4$\,$^\dagger$ & $47.1\%$ & $17$  & $+0.033$ & $52.9\%$ / $23.5\%$ & & $9$  & $+0.062$ & $11.1\%$ / $44.4\%$ \\
\bottomrule
\end{tabular}
\\[2pt]
\raggedright\footnotesize{$^\dagger$~Qwen3-VL-8B's $k\!\geq\!4$ bin contains only $n\!=\!17$ calls ($9$ non-saturated), versus $n\!=\!181/53/13$ at $k\!=\!1/2/3$. As discussed in \S\ref{sec:step_level_intervention}, the \emph{sustained per-call value} label is supported by the well-sampled region $k\!\leq\!3$, where the contrast with Mini-o3 is robust ($\sim\!0.5$ vs.\ $\sim\!0.04$ on the non-saturated subset at $k\!=\!3$). The apparent $k\!\geq\!4$ drop in this row ($+0.033$ all-calls, $+0.062$ non-sat) and the harmful-rate at $k\!\geq\!4$ ($23.5\%/44.4\%$) are reported for completeness but should be read as suggestive, not as decisive evidence of a tail collapse.}
\end{table*}

\subsection{POER vs.\ Saturation Rate}
\label{app:poer-vs-saturation}

Conceptually, POER (Eq.~\ref{eq:poer}) asks how often the policy keeps emitting tool calls after its confidence in the correct answer has already crossed the saturation threshold. It differs from the saturation rate $\Pr[g_{i-1}\!>\!\tau_{\text{sat}}]$ by jointly conditioning on $g_0\!\le\!\tau_{\text{sat}}$ and $k\!\geq\!2$: together, these two conditions restrict attention to trajectories that genuinely required tool use and ensure that any saturation we observe was \emph{caused by the policy's own earlier call(s)} rather than already present in the initial prompt.
Saturation rate describes \emph{when calls land in the saturated region}, whereas POER is a conditional certificate of \emph{whether the policy's stopping rule is calibrated against its own confidence}.

\section{Step-Level Intervention on Open-Ended Benchmarks}
\label{app:veg_openended}

The counterfactual intervention itself does not require a finite option set.
What is specific to MCQ is the bounded probability-gap readout
(Eq.~\ref{eq:prob_gap}) and the thresholded diagnostic derived from it.
We therefore extend the step-level analysis
(\S\ref{sec:step_level_intervention}) to the open-ended benchmark
\textbf{VisualProbe}~\citep{lai2025mini-o3} as supplementary evidence, while
retaining MCQ as the primary setting for the calibrated diagnostic.

For open-ended questions, the gold answer is a token sequence
$Y^\star=(y^\star_1,\ldots,y^\star_L)$ rather than one option label.
We retain the main-text notation and replace the MCQ probability gap $g_i$
with the following length-normalized local readout:
\begin{equation}
\label{eq:g-open}
g^{\text{open}}(C)
=\frac{1}{L}\sum_{t=1}^{L}\log P(y^\star_t\mid C,y^\star_{<t}).
\end{equation}
Without normalization, the joint log-likelihood is a sum over tokens and would
systematically favor short answers.
Averaging over $L$ instead measures the average support per gold token,
allowing answers of different lengths to be compared.
At token $t$, we condition on the preceding gold tokens $y^\star_{<t}$;
this evaluates the same target sequence in both branches.

For call $i$, let $C_i^-$ be the prefix before the call, and define
$C_i^{+,\text{real}}=(C_i^-,T_i,O_i^{\text{real}})$ and
$C_i^{+,\text{cf}}=(C_i^-,T_i,O_i^{\text{cf}})$.
As in the main text, we write
$g_{i-1}^{\text{open}}=g^{\text{open}}(C_i^-)$,
$g_i^{\text{real,open}}=g^{\text{open}}(C_i^{+,\text{real}})$,
and
$g_i^{\text{cf,open}}=g^{\text{open}}(C_i^{+,\text{cf}})$.
The per-step Marginal Gains are then
\begin{equation}
\label{eq:mg-open}
\begin{aligned}
\Delta M_i^{\text{real,open}}
&=g_i^{\text{real,open}}-g_{i-1}^{\text{open}},\\
\Delta M_i^{\text{cf,open}}
&=g_i^{\text{cf,open}}-g_{i-1}^{\text{open}}.
\end{aligned}
\end{equation}
The open-ended VEG has the same form as Eq.~\ref{eq:veg}:
\begin{equation}
\label{eq:veg-open}
\begin{aligned}
\mathrm{VEG}_i^{\text{open}}
&=\Delta M_i^{\text{real,open}}
-\Delta M_i^{\text{cf,open}}\\
&=g_i^{\text{real,open}}
-g_i^{\text{cf,open}}.
\end{aligned}
\end{equation}
The real and counterfactual branches have the same $C_i^-$ and $T_i$, and
differ only in the returned crop.
Thus, this is a local probe of whether the returned visual content increases
support for $Y^\star$, relative to a counterfactual crop.
It does not require a free-form answer rollout.
Because the open-ended VEG is on a different scale (nats per token, unbounded)
from the MCQ VEG (probability units, bounded in $[-2,2]$),
absolute magnitudes between this appendix and Table~\ref{tab:veg_results}
are not directly comparable; what the comparison preserves is the qualitative
pattern of observation-mediated influence.

\paragraph{Results on VisualProbe.}
We run this extension on VisualProbe for DeepEyes, Mini-o3, and Qwen3-VL-8B.
The score is in nats per gold token and has heavy tails, so we report the
median (IQR) over valid calls rather than the mean, using raw scores.

\begin{table*}[t]
\caption{\textbf{Open-ended $\mathrm{VEG}^{\text{open}}$ on VisualProbe.} Median (IQR) in nats per gold token, over all valid calls and split by trajectory correctness.}
\label{tab:app-veg-open}
\centering
\small
\setlength{\tabcolsep}{4pt}
\begin{tabular}{lccc}
\toprule
Model & All calls & Correct & Incorrect \\
\midrule
DeepEyes    & $+0.001$ ($0.909$) & $+0.001$ ($0.549$) & $-0.033$ ($1.098$) \\
Mini-o3     & $-0.027$ ($0.813$) & $+0.012$ ($0.656$) & $-0.165$ ($1.225$) \\
Qwen3-VL-8B & $+0.000$ ($2.424$) & $+0.017$ ($1.264$) & $-0.416$ ($4.317$) \\
\bottomrule
\end{tabular}
\end{table*}

The directional patterns in Table~\ref{tab:app-veg-open} match the MCQ results. DeepEyes has a near-zero central tendency and little separation between correct and incorrect trajectories, consistent with weak observation-mediated influence. Mini-o3 is more favorable on correct trajectories and negative on incorrect ones, consistent with useful evidence that is diluted or reversed on failed trajectories. Qwen3-VL-8B shows the largest correct-versus-incorrect separation, with a substantially negative incorrect median and a positive correct median, agreeing with the MCQ finding that useful and harmful observations coexist across its trajectories.

There are important differences. The open-ended score is unbounded and its IQR is wide, especially for Qwen3-VL-8B, so it does not support absolute-VEG comparisons across settings. The length-normalized gold likelihood is also not a margin against the strongest competing free-form answer. We therefore do not transfer the MCQ saturation threshold, near-zero cutoff, or four-group classifier to VisualProbe without separate calibration.

This distinction explains why MCQ remains preferable for the main diagnostic. The option-restricted softmax gives a common, bounded probability gap between the correct answer and its strongest distractor. It supports interpretable saturation and inactivity thresholds that are shared across calls and models. The open-ended extension demonstrates that the real-versus-counterfactual evidence contrast remains measurable beyond MCQ, but it is best viewed as supplementary evidence rather than a replacement for the MCQ diagnostic.

\section{Diagnostic Classification: Supporting Results}
\label{app:diagnostic-full}

This appendix supplies the supporting evidence for the trajectory-level
diagnostic of \S\ref{sec:diagnosing_policy_miscalibration}.
\S\ref{app:diagnostic-sensitivity} reports the joint sensitivity sweep over
the two continuous thresholds $(\tau_{\text{sat}}, \epsilon)$;
\S\ref{app:ate-decomposition} reports the group-wise decomposition
of each model's policy-level ATE; \S\ref{app:behavioral-validity} reports
the per-group behavioral audit (Mode~1 pre-call commitment and the four Mode~2
pathologies) on representative models; and \S\ref{app:hrbench-validation}
validates the four-group diagnostic and ATE decomposition on HR-Bench-4K.

\subsection{Diagnostic Threshold Sensitivity}
\label{app:diagnostic-sensitivity}

The main text fixes $\tau_{\text{sat}}\!=\!0.95$ and $\epsilon\!=\!0.01$.
We check whether the diagnostic conclusions depend on these choices in two
one-dimensional sweeps. Table~\ref{tab:app-diagnostic-eps-sensitivity} fixes
$\tau_{\text{sat}}\!=\!0.95$ and varies the near-zero VEG cutoff
$\epsilon$. Table~\ref{tab:app-diagnostic-tau-sensitivity} fixes
$\epsilon\!=\!0.01$ and varies the saturation cutoff $\tau_{\text{sat}}$.
For each setting we report all four group fractions, the Calibrated
contribution to the policy-level ATE, and the total policy-level ATE.

\begin{table*}[t]
\caption{\textbf{Sensitivity to the near-zero VEG threshold $\epsilon$ on
V$^{\!*}$.} $\tau_{\text{sat}}\!=\!0.95$ is fixed. ``Cal. contrib'' is
$f_{\text{Cal}}\bar\Delta_{\text{policy},\text{Cal}}$ in pp, and
$\mathrm{ATE}_{\text{policy}}$ is the row-sum contribution in pp.}
\label{tab:app-diagnostic-eps-sensitivity}
\centering
\small
\setlength{\tabcolsep}{3pt}
\begin{tabular}{lccccccc}
\toprule
Model & $\epsilon$ & No-call & Mode~1 & Cal. & Mode~2 & Cal. contrib & $\mathrm{ATE}_{\text{policy}}$ \\
\midrule
DeepEyes     & $0.005$ & $19.4$ & $72.3$ & $8.4$  & $0.0$ & $+0.5$ & $+0.0$ \\
DeepEyes     & $0.01$  & $19.4$ & $72.8$ & $7.9$  & $0.0$ & $+0.5$ & $+0.0$ \\
DeepEyes     & $0.02$  & $19.4$ & $73.8$ & $6.8$  & $0.0$ & $+0.0$ & $+0.0$ \\
DeepEyes     & $0.05$  & $19.4$ & $73.8$ & $6.8$  & $0.0$ & $+0.0$ & $+0.0$ \\
\addlinespace[2pt]
Mini-o3      & $0.005$ & $10.5$ & $29.8$ & $48.7$ & $11.0$ & $+3.5$ & $+5.5$ \\
Mini-o3      & $0.01$  & $10.5$ & $34.6$ & $45.0$ & $9.9$  & $+3.3$ & $+5.5$ \\
Mini-o3      & $0.02$  & $10.5$ & $38.2$ & $43.5$ & $7.9$  & $+3.7$ & $+5.5$ \\
Mini-o3      & $0.05$  & $10.5$ & $45.0$ & $39.3$ & $5.2$  & $+3.5$ & $+5.5$ \\
\addlinespace[2pt]
Qwen3-VL-8B  & $0.005$ & $5.2$  & $70.7$ & $20.9$ & $3.1$  & $+7.6$ & $+6.9$ \\
Qwen3-VL-8B  & $0.01$  & $5.2$  & $70.7$ & $20.9$ & $3.1$  & $+7.6$ & $+6.9$ \\
Qwen3-VL-8B  & $0.02$  & $5.2$  & $71.7$ & $20.4$ & $2.6$  & $+7.6$ & $+6.9$ \\
Qwen3-VL-8B  & $0.05$  & $5.2$  & $71.7$ & $20.4$ & $2.6$  & $+7.6$ & $+6.9$ \\
\addlinespace[2pt]
Qwen3-VL-4B  & $0.005$ & $13.1$ & $58.6$ & $24.1$ & $4.2$  & $+3.5$ & $+4.4$ \\
Qwen3-VL-4B  & $0.01$  & $13.1$ & $58.6$ & $24.1$ & $4.2$  & $+3.5$ & $+4.4$ \\
Qwen3-VL-4B  & $0.02$  & $13.1$ & $58.6$ & $24.1$ & $4.2$  & $+3.5$ & $+4.4$ \\
Qwen3-VL-4B  & $0.05$  & $13.1$ & $61.3$ & $23.0$ & $2.6$  & $+3.5$ & $+4.4$ \\
\bottomrule
\end{tabular}
\end{table*}

\begin{table*}[t]
\caption{\textbf{Sensitivity to the saturation threshold $\tau_{\text{sat}}$
on V$^{\!*}$.} $\epsilon\!=\!0.01$ is fixed. Columns follow
Table~\ref{tab:app-diagnostic-eps-sensitivity}.}
\label{tab:app-diagnostic-tau-sensitivity}
\centering
\small
\setlength{\tabcolsep}{3pt}
\begin{tabular}{lccccccc}
\toprule
Model & $\tau_{\text{sat}}$ & No-call & Mode~1 & Cal. & Mode~2 & Cal. contrib & $\mathrm{ATE}_{\text{policy}}$ \\
\midrule
DeepEyes     & $0.80$ & $19.4$ & $73.3$ & $7.3$  & $0.0$  & $+1.0$ & $+0.0$ \\
DeepEyes     & $0.90$ & $19.4$ & $72.8$ & $7.9$  & $0.0$  & $+0.5$ & $+0.0$ \\
DeepEyes     & $0.95$ & $19.4$ & $72.8$ & $7.9$  & $0.0$  & $+0.5$ & $+0.0$ \\
DeepEyes     & $0.97$ & $19.4$ & $72.3$ & $8.4$  & $0.0$  & $+0.5$ & $+0.0$ \\
DeepEyes     & $0.99$ & $19.4$ & $71.7$ & $8.9$  & $0.0$  & $+0.5$ & $+0.0$ \\
\addlinespace[2pt]
Mini-o3      & $0.80$ & $10.5$ & $43.5$ & $33.0$ & $13.1$ & $+2.0$ & $+5.5$ \\
Mini-o3      & $0.90$ & $10.5$ & $36.1$ & $41.4$ & $12.0$ & $+3.1$ & $+5.5$ \\
Mini-o3      & $0.95$ & $10.5$ & $34.6$ & $45.0$ & $9.9$  & $+3.3$ & $+5.5$ \\
Mini-o3      & $0.97$ & $10.5$ & $32.5$ & $48.2$ & $8.9$  & $+3.4$ & $+5.5$ \\
Mini-o3      & $0.99$ & $10.5$ & $32.5$ & $52.4$ & $4.7$  & $+3.5$ & $+5.5$ \\
\addlinespace[2pt]
Qwen3-VL-8B  & $0.80$ & $5.2$  & $72.3$ & $18.3$ & $4.2$  & $+7.4$ & $+6.9$ \\
Qwen3-VL-8B  & $0.90$ & $5.2$  & $70.7$ & $19.9$ & $4.2$  & $+7.6$ & $+6.9$ \\
Qwen3-VL-8B  & $0.95$ & $5.2$  & $70.7$ & $20.9$ & $3.1$  & $+7.6$ & $+6.9$ \\
Qwen3-VL-8B  & $0.97$ & $5.2$  & $69.6$ & $21.5$ & $3.7$  & $+7.6$ & $+6.9$ \\
Qwen3-VL-8B  & $0.99$ & $5.2$  & $66.5$ & $24.6$ & $3.7$  & $+7.8$ & $+6.9$ \\
\addlinespace[2pt]
Qwen3-VL-4B  & $0.80$ & $13.1$ & $62.8$ & $20.4$ & $3.7$  & $+3.7$ & $+4.4$ \\
Qwen3-VL-4B  & $0.90$ & $13.1$ & $61.3$ & $22.5$ & $3.1$  & $+3.8$ & $+4.4$ \\
Qwen3-VL-4B  & $0.95$ & $13.1$ & $58.6$ & $24.1$ & $4.2$  & $+3.5$ & $+4.4$ \\
Qwen3-VL-4B  & $0.97$ & $13.1$ & $56.0$ & $26.2$ & $4.7$  & $+3.5$ & $+4.4$ \\
Qwen3-VL-4B  & $0.99$ & $13.1$ & $54.5$ & $28.8$ & $3.7$  & $+3.5$ & $+4.4$ \\
\bottomrule
\end{tabular}
\end{table*}

These sweeps show that the main conclusion is not an artifact of one chosen threshold. DeepEyes remains
Mode~1-dominant with no Mode~2 items; Qwen3-VL-8B and Qwen3-VL-4B remain
Mode~1-dominant with non-trivial Calibrated tails; Mini-o3 remains the most
Calibrated-heavy model. The ATE localization is also stable: the Calibrated
contribution remains positive and accounts for most of the positive
policy-level ATE for Mini-o3, Qwen3-VL-8B, and Qwen3-VL-4B throughout both sweeps.

There is only one expected boundary case: Mini-o3 at $\tau_{\text{sat}}=0.80$. This cutoff is a lenient criterion for declaring saturation, since a probability gap of $0.80$ can still leave meaningful uncertainty and room for visual evidence to help. As a result, some trajectories that are useful under the stricter high-confidence cutoffs are reclassified as Mode 1 or Mode 2. Even in this boundary case, the Calibrated subset remains positive ($+2.0$ pp), and for the more natural high-confidence thresholds $\tau_{\text{sat}}\ge 0.90$, the Calibrated contribution is again the largest source of Mini-o3's positive ATE. Thus the qualitative localization of gains to useful visual-evidence trajectories is stable under reasonable threshold choices.

\subsection{Group Decomposition of the Policy-Level ATE}
\label{app:ate-decomposition}

For each model the policy-level ATE ($\text{ATE}_{\text{policy}}$) certified in
\S\ref{sec:policy_level_intervention} can be decomposed according to the group distribution
\begin{equation}
\label{eq:ate-decomposition}
\text{ATE}_{\text{policy}} \;=\; \sum_{b\in\mathcal{B}} f_b \cdot \bar\Delta_{\text{policy},b},
\end{equation}
where
$\mathcal{B}\!=\!\{\text{No-call},\text{Mode 1},\text{Calibrated},\text{Mode 2}\}$,
$f_b$ is the fraction of group $b$ on V$^{\!*}$, and $\bar\Delta_{\text{policy},b}$
is the mean accuracy gap between $\pi_{\text{tool}}$ and $\pi_{\text{direct}}$ on the trajectories in group $b$.

Table~\ref{tab:ate_decomp_full} reports the full three-line breakdown
($f_b$, $\bar\Delta_{\text{policy},b}$, and the resulting contribution
$f_b\!\cdot\!\bar\Delta_{\text{policy},b}$) underlying the compact
Table~\ref{tab:ate_decomp} of the main text. The rightmost Total is computed from unrounded contributions and matches the policy-level ATE in Table~\ref{tab:policy_ate}; displayed components may not sum exactly due to rounding.

\begin{table*}[t]
\caption{\textbf{Per-model group decomposition of the policy-level ATE on
V$^{\!*}$.} For each model: $f_b$ is the fraction of group $b$ (\%);
$\bar\Delta_{\text{policy},b}$ is the mean accuracy gap between $\pi_{\text{tool}}$ and $\pi_{\text{direct}}$ on the trajectories in group $b$ (pp); contrib $=f_b\!\cdot\!\bar\Delta_{\text{policy},b}$ (pp).
Total is from unrounded contributions and may differ slightly from the sum of displayed cells due to rounding.
``--'' marks an empty group on this benchmark.}
\label{tab:ate_decomp_full}
\centering
\small
\setlength{\tabcolsep}{5pt}
\begin{tabular}{llrrrrr}
\toprule
Model & & No-call & Mode 1 & Cal. & Mode 2 & Total \\
\midrule
DeepEyes
 & $f_b$ (\%)            & $19.4$ & $72.8$ & $7.9$  & $0.0$  & $100.0$ \\
 & $\bar\Delta_{\text{policy},b}$ (pp) & $0.0$  & $-0.7$ & $+6.7$ & $-$    & --      \\
 & contrib (pp)          & $0.0$  & $-0.5$ & $+0.5$ & $0.0$  & $0.0$   \\
\addlinespace[2pt]
Qwen3-VL-4B
 & $f_b$ (\%)            & $13.1$ & $58.6$ & $24.1$ & $4.2$  & $100.0$ \\
 & $\bar\Delta_{\text{policy},b}$ (pp) & $-1.5$ & $+3.3$ & $+14.4$ & $-20.3$ & --     \\
 & contrib (pp)          & $-0.2$ & $+2.0$ & $+3.5$ & $-0.9$ & $+4.4$  \\
\addlinespace[2pt]
Qwen3-VL-8B
 & $f_b$ (\%)            & $5.2$  & $70.7$ & $20.9$ & $3.1$  & $100.0$ \\
 & $\bar\Delta_{\text{policy},b}$ (pp) & $-8.8$ & $-0.3$ & $+36.2$ & $-2.1$  & --     \\
 & contrib (pp)          & $-0.5$ & $-0.2$ & $+7.6$ & $-0.1$ & $+6.9$  \\
\addlinespace[2pt]
Mini-o3
 & $f_b$ (\%)            & $10.5$ & $34.6$ & $45.0$ & $9.9$  & $100.0$ \\
 & $\bar\Delta_{\text{policy},b}$ (pp) & $-2.5$ & $+5.5$ & $+7.3$ & $+5.9$  & --     \\
 & contrib (pp)          & $-0.3$ & $+1.9$ & $+3.3$ & $+0.6$ & $+5.5$  \\
\bottomrule
\end{tabular}
\end{table*}

\subsection{Behavioral Validity for Group Labels}
\label{app:behavioral-validity}

The diagnostic of \S\ref{sec:diagnosing_policy_miscalibration} is structural;
this appendix tests its labels against \emph{independent} behavioral
signatures that the per-call features do not directly encode. Here we provide more evidence.

\paragraph{Mode~1 trajectories: performative calls.}
For each Mode~1 trajectory we extract the first explicit answer expression in
the pre-call $\texttt{<think>}$ block (e.g., ``\emph{the answer
should be X}'', ``\emph{I think it's X}'') and compare it against the final answer
$Y$. Across DeepEyes, Qwen3-VL-8B, and Mini-o3 on
V$^{\!*}$, the $\texttt{<think>}$ blocks already commit to the final $Y$ in $\sim\!58\%$ of Mode~1 trajectories, compared with $\sim\!28\%$ for Calibrated trajectories. Mode~1 trajectories thus read as \emph{performative}: visual
tool-use without visual perception, with the call serving as a downstream
by-product of an already-committed prediction rather than a mechanism for
revising belief.

\paragraph{Mode~2 trajectories: four pathologies of failed integration.}
We audit $1{,}528$ Mini-o3 rollouts on V$^{\!*}$ ($8\!\times\!191$ instances)
which contribute the bulk of our Mode~2 trajectories, and find four
pathologies that jointly characterize the group label.
\textbf{(P1)~Lower accuracy with more calls:} accuracy peaks at $N\!=\!3$ calls
($78.2\%$) and collapses to $0\%$ at the $N\!=\!12$ limit.
\textbf{(P2)~Local fixation:} $38\%$ of crops occupy $<\!4\%$ of the source
image and failed trajectories repeatedly issue $\pm 3\%$ adjustments
\emph{within the same window} rather than relocating---no deployment of
standard search primitives.
\textbf{(P3)~Non-progressive restarts:} on the budget-saturated subset, the
fraction of restarts (vs.\ hierarchical refinement) doubles from $27.6\%$ to
$56.4\%$, indicating uncoordinated guessing precisely where coherent zoom is
most needed.
\textbf{(P4)~Absent deadline-aware commitment:} of the $20$
trajectories that hit the tool-call limit, all $20$ are wrong and \emph{none} commits a final answer at the last step.

\subsection{Cross-Benchmark Validation on HR-Bench-4K}
\label{app:hrbench-validation}

To check that the four-group diagnostic and the ATE decomposition are not specific to V$^{\!*}$, we apply the same step-level VEG protocol and the same deterministic diagnostic rule ($\tau_{\text{sat}}\!=\!0.95$, $\epsilon\!=\!0.01$) to HR-Bench-4K, using DeepEyes, Mini-o3, and Qwen3-VL-8B on $800$ items.

\setcounter{dbltopnumber}{3}

\begin{table*}[t]
\caption{\textbf{Step-level VEG on HR-Bench-4K.} \emph{VEG C / I} splits by trajectory correctness; \emph{Sat-rate} is the fraction of saturated calls; \emph{mean $V^{\max}$} is the per-trajectory best call averaged over trajectories.}
\label{tab:app-hrbench-veg}
\centering
\small
\setlength{\tabcolsep}{5pt}
\begin{tabular}{lccccccc}
\toprule
Model & Valid calls & VEG all & VEG C / I & $\lvert\mathrm{VEG}\rvert\!<\!0.01$ & Sat-rate & mean $V^{\max}$ & Pattern \\
\midrule
DeepEyes    & $769$  & $-0.011$ & $+0.004$ / $-0.048$ & $86.9\%$ & $34.9\%$ & $-0.005$ & Mode-1-heavy \\
Mini-o3     & $2398$ & $+0.030$ & $+0.115$ / $-0.023$ & $25.4\%$ & $12.6\%$ & $+0.171$ & Calibrated-heavy, Mode~2 \\
Qwen3-VL-8B & $1049$ & $+0.036$ & $+0.099$ / $-0.150$ & $64.6\%$ & $58.3\%$ & $+0.143$ & Saturation-driven Mode~1 \\
\bottomrule
\end{tabular}
\end{table*}

\begin{table*}[t]
\caption{\textbf{Group distribution on HR-Bench-4K.} Rows sum to $100\%$; $(\tau_{\text{sat}}, \epsilon)\!=\!(0.95, 0.01)$.}
\label{tab:app-hrbench-groups}
\centering
\footnotesize
\setlength{\tabcolsep}{4pt}
\begin{tabular}{lcccc}
\toprule
Model & No-call & Mode 1 (CWL) & Calibrated & Mode 2 (LWP) \\
\midrule
DeepEyes    & $9.5$  & $85.6$ & $4.9$  & $0.0$ \\
Mini-o3     & $16.1$ & $25.8$ & $42.8$ & $15.4$ \\
Qwen3-VL-8B & $29.8$ & $58.4$ & $10.1$ & $1.6$ \\
\bottomrule
\end{tabular}
\end{table*}

\begin{table*}[t]
\caption{\textbf{Group-wise ATE decomposition on HR-Bench-4K} (pp). Rows sum to the total ATE.}
\label{tab:app-hrbench-ate}
\centering
\footnotesize
\setlength{\tabcolsep}{3pt}
\begin{tabular}{lccccc}
\toprule
Model & No-call & Mode 1 & Cal. & Mode 2 & Total \\
\midrule
DeepEyes    & $-0.5$ & $-1.5$ & $+0.6$ & $0.0$  & $-1.4$ \\
Mini-o3     & $+0.5$ & $+0.5$ & $+3.4$ & $+0.4$ & $+4.8$ \\
Qwen3-VL-8B & $+1.4$ & $-0.9$ & $+4.4$ & $+0.1$ & $+5.0$ \\
\bottomrule
\end{tabular}
\end{table*}

The HR-Bench-4K results reproduce the same qualitative signatures (Table~\ref{tab:app-hrbench-veg}). DeepEyes stays largely inactive, with $86.9\%$ of calls near-zero and a per-trajectory best near zero. Mini-o3 stays information-driven but diluted, with a clear correct-versus-incorrect gap ($+0.115$ vs.\ $-0.023$) and a positive best call ($+0.171$). Qwen3-VL-8B stays saturation-dependent, with $58.3\%$ saturated calls that carry near-zero VEG while non-saturated calls remain positive.

The group distribution (Table~\ref{tab:app-hrbench-groups}) and the ATE decomposition (Table~\ref{tab:app-hrbench-ate}) support the same high-level conclusion beyond V$^{\!*}$: the Calibrated subset is positive for all three models and is the largest contributor for Mini-o3 and Qwen3-VL-8B. For DeepEyes the overall negative ATE is driven by Mode~1 and No-call, whose negative contributions outweigh the small positive Calibrated term. The grouping is therefore not a V$^{\!*}$-only artifact.

\section{Declaration of LLM usage}
\label{app:llm_usage}

The authors used large language models (LLMs) only for polishing prose of text where the complete draft was fully written by the authors initially and polished later with the help of LLM-based assistants including ChatGPT and Gemini. The authors used code assistant Cursor to implement the authors' original design and ideas. The scientific contributions, technical methods, ideas and core results are entirely the original work of the authors.

\end{document}